\documentclass[sigconf]{acmart}
\usepackage{amsmath,amsfonts}
\usepackage{array}
\usepackage[caption=false,font=normalsize,labelfont=sf,textfont=sf]{subfig}
\usepackage{textcomp}
\usepackage{stfloats}
\usepackage{url}
\usepackage{verbatim}
\usepackage{graphicx}
\usepackage{paralist}
\usepackage{amsmath,amsfonts}
\usepackage{array}
\usepackage[caption=false,font=normalsize,labelfont=sf,textfont=sf]{subfig}
\usepackage{textcomp}
\usepackage{stfloats}
\usepackage{url}
\usepackage{verbatim}
\usepackage{graphicx}
\usepackage[ruled,vlined,linesnumbered]{algorithm2e}
\usepackage{algpseudocode}
\usepackage{xcolor}
\usepackage{multirow}
\usepackage{amsfonts}
\usepackage{xspace}
\usepackage{colortbl}
\usepackage{makecell}
\usepackage{microtype}
\usepackage{enumitem}
\usepackage{pifont}
\usepackage{booktabs}
\usepackage[switch]{lineno}

\newcommand{\tool}{\textit{ADReFT}\xspace}
\newcommand{\encoder}{\textit{State Encoder}\xspace}
\newcommand{\monitor}{\textit{State Monitor}\xspace}
\newcommand{\repair}{\textit{Decision Adapter}\xspace}

\AtBeginDocument{%
  }

\setcopyright{none}
\renewcommand\footnotetextcopyrightpermission[1]{}
\copyrightyear{2018}
\acmYear{2018}
\acmDOI{XXXXXXX.XXXXXXX}
\acmConference[Conference acronym 'XX]{Make sure to enter the correct
  conference title from your rights confirmation email}{June 03--05,
  2018}{Woodstock, NY}
\acmISBN{978-1-4503-XXXX-X/2018/06}

\begin{document}

\title{\tool: Adaptive Decision Repair for Safe Autonomous Driving via Reinforcement Fine-Tuning}

\author{Mingfei Cheng}
\affiliation{
  \institution{Singapore Management University}
  \country{Singapore}
}

\author{Xiaofei Xie}
\affiliation{
  \institution{Singapore Management University}
  \country{Singapore}
}

\author{Renzhi Wang}
\affiliation{
  \institution{University of Alberta}
  \country{Canada}
}

\author{Yuan Zhou}
\affiliation{
  \institution{Zhejiang Sci-Tech University}
  \country{China}
}

\author{Ming Hu}
\affiliation{
  \institution{Singapore Management University}
  \country{Singapore}
}

\renewcommand{\shortauthors}{Mingfei Cheng, Xiaofei Xie, Renzhi Wang, Yuan Zhou and Ming Hu}

\renewcommand{\shortauthors}{Trovato et al.}

\begin{abstract}
%

%
Autonomous Driving Systems (ADSs)
continue to face safety-critical risks due to the inherent limitations in their design and performance capabilities.
Online repair plays a crucial role in mitigating such limitations, ensuring the runtime safety and reliability of ADSs.
Existing online repair solutions enforce ADS compliance by transforming unacceptable trajectories into acceptable ones based on predefined specifications, such as rule-based constraints or training datasets.
However, these approaches often lack generalizability, adaptability and tend to be overly conservative, resulting in ineffective repairs that not only fail to mitigate safety risks sufficiently but also degrade the overall driving experience.

To address this issue, we propose Adaptive Decision Repair (\tool), a novel and effective repair method that identifies safety-critical states through offline learning from failed tests and generates appropriate mitigation actions to improve ADS safety.
Specifically, \tool{} incorporates a transformer-based model with two joint heads, \monitor{} and \repair{}, designed to capture complex driving environment interactions to evaluate state safety severity and generate adaptive repair actions.
Given the absence of oracles for state safety identification, we first pretrain \tool{} using supervised learning with coarse annotations, i.e., labeling states preceding violations as positive samples and others as negative samples.
It establishes \tool{}'s foundational capability to mitigate safety-critical violations, though it may result in somewhat conservative mitigation strategies.
Therefore, we subsequently fine-tune \tool{} using reinforcement learning to improve its initial capability and generate more precise and contextually appropriate repair decisions.

We evaluated \tool on two leading ADSs within the CARLA simulator, including an end-to-end ADS \textit{Roach} and a module-based ADS \textit{Pylot}.
Extensive experimental results demonstrate that \tool{} effectively mitigates safety-critical collisions, improving safety by an average of 80.5\% and outperforming the best baseline by 32.8\%. Morevoer, \tool{} achieves better repair performance by introducing fewer interventions on the original ADS decisions, with an average intervention intensity of only 0.75, outperforming the best baseline by 25.7\%. 
\end{abstract}

\maketitle

\section{Introduction}
Autonomous Driving Systems (ADSs) are revolutionizing transportation and urban mobility by enabling vehicles to navigate and make real-time driving decisions. 
These systems rely on a variety of sensors, such as cameras and LiDAR, to perceive their surroundings and interpret the driving environment.
The decision-making process in ADSs is primarily driven by two complementary approaches: predefined rule-based strategies derived from expert knowledge and data-driven artificial intelligence (AI) models. 
While recent advancements in AI have significantly enhanced ADS capabilities, accelerating their real-world adoption, these methods still face safety-critical challenges in generalizing effectively across diverse and dynamic driving conditions~\cite{garcia2020comprehensive,zhou2023specification}.

Extensive ADS testing techniques~\cite{av_fuzzer, cheng2023behavexplor, icse_samota, tse_adfuzz} have been widely proposed to evaluate the Safety of the Intended Functionality (SOTIF) of ADSs by generating safety-critical scenarios that expose potential issues, such as collisions. 
However, since existing ADSs remain susceptible to even small perturbations, these approaches often generate a large number of failure cases. Hence, a challenge is how to effectively and efficiently leveraging these safety-critical scenarios to enhance ADS safety.
To tackle this challenge, a common approach is Offline Repair~\cite{monperrus2018automatic}, where developers invest significant effort in identifying the root causes of failure cases and iteratively crafting repair patches (e.g., retraining models) to address the issues within the development environment. However, offline repair is often time-consuming and limited in scope, as it cannot be applied once the system is deployed (e.g., in sold vehicles).


In contrast, \textit{Online Repair}~\cite{monperrus2018automatic}, which involves performing repairs on deployed software through dynamic software updates (DSU), offers a promising approach to enhancing ADSs without necessitating full system-level reengineering.
Existing ADS online repair techniques typically falls into two categories: \textit{Rule-based Enforcement}~\cite{mauritz2016assuring, watanabe2018runtime, shankar2020formal, grieser2020assuring, hong2020avguardian, sun2024redriver, sun2025fixdrive} and \textit{Learning-based Methods}~\cite{chen2019deep, grieser2020assuring, stocco2020misbehaviour, stocco2022thirdeye}.
The first category, \textit{Rule-based Enforcement}, relies on predefined rule-based specifications (e.g., minimum safe distance from surrounding obstacles) to detect unexpected states and apply corrective actions, ensuring the ADS consistently adheres to the specified rules. 
However, these rules heavily rely on expert knowledge tailored to specific scenarios, limiting their applicability across diverse driving conditions. In particular, newly emerging safety-critical scenarios still require significant effort to design corresponding specifications.  
For instance, specifications developed for highway scenarios may require substantial human effort to be adapted for city intersections, where traffic dynamics and constraints differ considerably.
The second category, \textit{Learning-based Methods}, derives safety specifications from normal driving data (i.e., safe scenarios observed during testing) to identify and rectify abnormal states, offering a more flexible solution.  
Anomaly-based methods are often prone to false positives due to the dependence on limited normal driving data, necessitating a highly adaptive repair strategy.
However, existing works often heuristically apply emergency braking in response to detected dangerous situations. As a result, such approaches often lead to frequent yet ineffective interventions that degrade driving performance. 
Therefore, there is a need for an online repair approach that can not only effectively identify safety-critical states but also adaptively make repair decisions based on the detected driving states.

An ideal repair should select repair decisions based on the severity of the identified safety-critical driving state, applying minimal intervention when the situation is less critical. 
There are two main technical challenges for such a repair:  
\ding{182} The first challenge is effectively identifying safety-critical states before violations occur.  
The driving environment consists of numerous vehicles with intricate interactions, where violations often arise from minor perturbations induced by these interactions. However, no explicit oracles or specifications exist to reliably determine which interactions may lead to safety-critical violations. For example, rule-based approaches require precisely tuned detection hyperparameters to adapt to different scenarios, making them impractical for generalization.  
Thus, a major challenge lies in accurately modeling these complex interactions and proactively identifying potential safety-critical states before violations occur.  
\ding{183} The second challenge is determining an adaptive repair decision for identified safety-critical states.  
Specifically, the vast decision space in driving scenarios allows multiple strategies to avoid violations, leading to various repair solutions with inherent trade-offs.  
Arbitrarily selecting a repair action without considering the driving context may introduce unintended negative effects.  
For example, heuristically applying hard braking may cause excessive and disruptive interventions, degrading overall driving performance if safety-critical states are frequently identified. Conversely, mild interventions may be insufficient to effectively mitigate violations.  
Thus, designing a repair strategy that adaptively adjusts interventions to ensure safety while minimizing the impact on driving experience remains a challenge.



To address these challenges, we propose an Adaptive Decision Repair framework, \tool, which leverages a designed Reinforcement Fine-Tuning (ReFT) approach to automatically learn adaptive repair decisions from testing suites.  
Specifically, \tool{} consists of three modules: \encoder{}, which encodes driving states by capturing interactions between the ego vehicle and its surroundings; \monitor{}, which evaluates the safety of the encoded driving state; and \repair{}, which adaptively selects repair decisions for identified safety-critical states.  
To address the first challenge \ding{182}, we adopt a transformer-based encoder in \encoder{}, which processes object-level attributes (e.g., location, speed) to capture interactions among traffic participants using the attention mechanism.  
To enable \tool{} to effectively learn to identify safety-critical states, we first assign weak annotations to each driving state in the testing cases, labeling states preceding violations as positive samples and others as negative samples. \tool{} is then trained on these weakly annotated states using supervised learning (SL) to develop a fundamental ability to identify safety-critical states while initially prioritizing the most conservative repair decisions (i.e., hard braking) to ensure safety.  
To tackle the second challenge \ding{183}, we draw inspiration from Reinforcement Learning (RL) and further fine-tune only \repair{} in \tool{} using an online RL algorithm, denoted as ReFT. This enables \repair{} to explore more adaptive and optimal repair decisions based on the safe decisions learned in the previous stage.  
Specifically, in this stage, we introduce a \textit{safe-explore} reward derived from \tool{}, combined with a termination reward from scenario execution results, to guide the learning process. 
The ReFT stage ultimately enables \tool{} to adaptively adjust repair decisions based on current driving states, resulting in more effective repairs with minimal disruption to original driving stability.  

We evaluate \tool{} on two different types of ADSs: an end-to-end ADS, \textit{Roach}~\cite{roach_iccv}, and a module-based ADS, \textit{Pylot}~\cite{gog2021pylot}, across five scenarios in the CARLA simulator~\cite{dosovitskiy2017carla}, with a total of 800 testing cases (400 collision cases and 400 successful cases).  
The evaluation results demonstrate the effectiveness of \tool{} in mitigating safety-critical collisions. Specifically, \tool{} successfully repairs an average of 85\% of collisions for \textit{Roach} and 76\% for \textit{Pylot}, surpassing the best baseline, \textit{RuleRepair}, which relies on rule-based specifications for runtime intervention, by 30\% and 36\%, respectively.
Moreover, compared to the best baseline, \tool{} achieves the lowest intervention intensity in modifying ADS decisions, with an average intensity of only 0.77 for \textit{Roach} and 0.72 for \textit{Pylot}, significantly lower than the best baseline's 1.11 and 0.91, respectively—an improvement of 25.7\%.  
Additional experiments further validate the effectiveness of \monitor{} and \repair{} training strategies, highlighting their contributions to \tool{}'s performance.  
In summary, this paper makes the following contributions:
\begin{enumerate}[leftmargin=*]

\item We propose \tool{}, an adaptive online repair approach that identifies safety-critical states and adaptively performs repair based on the driving state while minimizing intervention intensity for the ADS.  


\item We design an automated training pipeline that combines supervised learning and reinforcement fine-tuning, enabling \tool{} to explore safe and adaptive repair decisions from test cases, bridging the gap between testing and automatic ADS repair.


\item Extensive experiments across various ADSs and driving scenarios demonstrate the effectiveness of \tool{} in repairing violations while minimizing intervention intensity, outperforming the best baseline by 32.8\% and 25.7\%, respectively.  
\end{enumerate}

\section{Background}

\subsection{Autonomous Driving System}\label{sec: background_ads}

Autonomous Driving Systems (ADSs) function as the core intelligence of autonomous vehicles, governing their behavior and decision-making. Existing ADSs can be categorized into two classes: \textit{Module-based} and \textit{End-to-End (E2E)}.

(1) \textit{Module-based ADS.} Module-based ADSs, such as Pylot~\cite{gog2021pylot}, Apollo~\cite{apollo}, and Autoware~\cite{autoware}, typically consist of multiple interconnected modules, including perception, prediction, planning, and control, which work together to generate driving actions.
The perception module leverages deep learning to detect surrounding objects, such as vehicles and pedestrians. The prediction module then tracks these objects and forecasts their motion trajectories. Using this information, the planning module generates a collision-free trajectory for the ego vehicle. Finally, the control module converts the planned trajectory into vehicle control commands—steering, throttle, and braking. 
However, the complex asynchronous communication mechanisms in module-based ADSs (e.g., CyberRT) often lead to non-deterministic behavior, making it challenging to verify the effectiveness of runtime intervention approaches by rerunning violation scenarios to repair failures (i.e., collision scenarios).  
Therefore, we deliberately choose Pylot~\cite{gog2021pylot} for our experiments, as its synchronized mode reduces non-deterministic behavior, making it particularly suitable for verifying this task.

(2) \textit{End-to-End ADS.} Recent advances in deep learning have shown significant progress in autonomous driving. End-to-End (E2E) based ADSs (i.e., Openpilot~\cite{openpilot}) treat the separate modules of \textit{Module-based} ADSs as a unified neural network. 
This network takes sensor data as input and outputs the final control commands to operate the vehicle. 
E2E ADSs~\cite{renz2023plant,hu2023_uniad,openpilot,roach_iccv} typically require a substantial amount of data to achieve satisfactory performance. For instance, it is estimated that Tesla's Full Self-Driving (FSD)~\cite{tesla_fsd} system has been trained on approximately 6 billion miles of data. 
Consequently, some approaches~\cite{roach_iccv, renz2023plant} focus solely on training the decision components, including prediction, planning, and control, in an end-to-end manner, utilizing perception results from pre-trained backbones such as BEVFusion~\cite{liu2023bevfusion}. 
In this paper, we select Roach~\cite{roach_iccv}, a state-of-the-art end-to-end (E2E) ADS designed to prioritize decision-making capabilities within ADSs.

Note that in this paper, we focus on the online repair of decision errors in ADSs to mitigate high-risk behaviors, explicitly targeting violations arising from the decision-making modules while excluding those caused by perception inaccuracies. 
Accordingly, all ADSs evaluated in our study operate with perfect (noise-free) perception outputs provided by the simulator to generate final control commands. 
This setup ensures that the effectiveness of online repair is assessed exclusively with respect to decision-making quality, isolating it from the influence of perception errors.

\subsection{Reinforcement Learning}\label{sec:background_RL}
Reinforcement Learning (RL) is a machine learning technique that allows an RL agent to learn through trial and error by interacting with an environment and receiving feedback based on its actions and observations. 
The behavior of a RL agent can be modeled as a Markov Decision Process (MDP), defined as $\mathcal{M} = (\mathcal{S}, \mathcal{A}, \mathcal{P}, \mathcal{R})$, where $\mathcal{S}$ denotes the state space, $\mathcal{A}$ denotes the action space, $\mathcal{P}: \mathcal{S} \times \mathcal{A} \times \mathcal{S} \rightarrow [0,1]$ represents the state transition probability function, and $\mathcal{R}: \mathcal{S} \times \mathcal{A} \rightarrow \mathbb{R}$ is an immediate reward function that calculates the reward for an action $a \in \mathcal{A}$ taken in a state $s \in \mathcal{S}$. 
In detail, at each time step $t$, the agent observes the state $s_{t}$ from the environment and selects an action $a_{t}$ from the action space $\mathcal{A}$, based on its policy $\pi(a_{t}|s_{t})$. This policy maps the observed state $s_t$ to the action $a_t$. Concurrently, the system assigns a reward $r_{t}=\mathcal{R}(s_t,a_t)$, indicating the effectiveness of the action $a_{t}$ in the state $s_{t}$ towards achieving the final goal (e.g., reaching a destination without collisions). The environment then transitions to the next state $s_{t+1}$. This process repeats until a terminal state is reached (e.g., success or failure). Throughout these interactions, the agent aims to maximize the cumulative future rewards in any state $R_{t} = \sum_{k=0}^{\infty}\gamma^{k} r_{t+k}$, where $\gamma \in [0,1]$ is a discount factor that balances short-term and long-term rewards.
In this work, we define the repair action space as a discrete set of actions and adopt Deep Q-Network (DQN) as the underlying RL algorithm due to its simplicity and theoretical guarantees of convergence to an optimal policy~\cite{wu2024recent}. 

\textbf{Deep Q-Network.} 
DQN extends traditional Q-learning by employing a deep neural network to approximate the Q-function $\mathcal{Q}_{\theta}(s, a)$, where $\theta$ denotes the learnable parameters of the network. 
The Q-function is updated by:
\begin{equation}\label{eq:DQN_Q} 
\mathcal{Q}_{\theta}(s, a) = \mathcal{R}(s, a) + \gamma \max_{a'} \mathcal{Q}_{\hat{\theta}}(s', a')
\end{equation}
where $\mathcal{R}(s, a)$ denotes the immediate reward for executing action $a$ in state $s$, and $\gamma$ is the discount factor that balances future rewards. 
The term $\mathcal{Q}_{\hat{\theta}}(s', a')$ denotes the target Q-value, computed using a separate target network with parameters $\hat{\theta}$ to improve training stability. 
The network parameters $\theta$ are optimized by minimizing the following loss function:
\begin{equation}\label{eq:DQN_Loss} 
L(\theta) = \mathbb{E}_{(s, a, r, s') \sim \mathcal{D}} \left[ \left( r + \gamma \max_{a'} \mathcal{Q}_{\hat{\theta}}(s', a') - \mathcal{Q}_{\theta}(s, a) \right)^2 \right] 
\end{equation} 
where $\mathcal{D}$ denotes the experience replay buffer, which stores past transitions $(s, a, r, s')$ and samples them uniformly at random to decorrelate updates and improve data efficiency.

\subsection{Scenario}
In the context of autonomous driving, a \textit{scenario} specifies the driving environment (e.g., road, weather, and illumination) along with the scenery and objects (i.e., Non-Player Character (NPC) vehicles and pedestrians). A \textit{scenario} can be described as a set $S = \{\mathbb{A}, \mathbb{P}, \mathbb{E}\}$. $\mathbb{A}$ denotes the motion task of the ADS under test, including the start position and destination. $\mathbb{P}$ is a finite set of participants, such as dynamic NPC vehicles. And $\mathbb{E}$ represents other environmental factors (e.g., weather). Typically, in the simulation environment~\cite{av_fuzzer, cheng2023behavexplor, icse_samota}, given a scenario $S$, the simulator executes the scenario by firstly parsing the configuration. 
At each time $t$, the simulator returns the \textit{scene observation} $\mathbf{o}_t = \{o_t^0, o_t^1, \ldots, 0_t^{|\mathbb{P}|}\}$ where $o_t^j = \{x_{t}^{j}, y_{t}^{j}, \theta_{t}^{j}, v_{t}^{j}\}$ represents the waypoint of participant $j \in \mathbb{P}$ at timestamp $t$, including the position $(x_{t}^{j}, y_{t}^{j})$, the heading $\theta_{t}^{j}$, and the velocity $v_{t}^{j}$. Once the ego vehicle reaches the destination or triggers violations (i.e., collision), the simulator terminates the execution of the scenario and returns the \textit{scenario result}, which indicates whether the ADS completes the scenario $S$.

\begin{figure}[!t]
    \centering
    \includegraphics[width=1.0\linewidth]{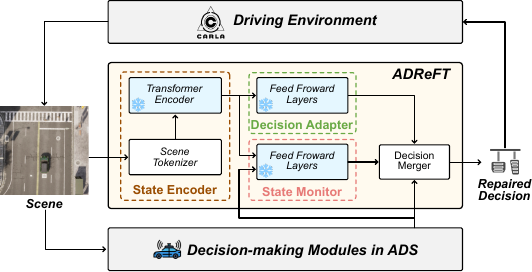}
    \caption{Architecture overview of \tool.}
    \vspace{-5pt}
    \label{fig:method_overview}
\end{figure}
\section{Approach}

The following sections introduce the architecture of \tool and provide a detailed description of its training process, encompassing both the Supervised Training Stage and the Reinforcement Fine-tuning Stage to achieve a satisfactory \tool{}.

\subsection{Architecture of \tool}

Figure~\ref{fig:method_overview}(a) presents the architecture of \tool{}, which functions as a plug-in module for the ADS under repair. Specifically, \tool{} consists of three main components: \encoder, \monitor{}, and \repair{}.
During runtime deployment, \tool{} takes as input both the driving scene and the corresponding ADS-generated decision, collectively referred to as the driving state in the following content.  
The driving state is first processed by the \textit{State Encoder}, which captures high-level feature representations of the surrounding environment (e.g., interactions between vehicles). 
The extracted environmental features, together with the original ADS decision, are then passed to the \monitor{} to predict a safety-critical score. Simultaneously, the \repair{} predicts an appropriate repair action for the current driving state.
Finally, a rule-based \textit{Decision Merger} generates the final driving decision by integrating the predicted safety-critical score, the predicted repair action, and the original ADS decision.
The following subsections present detailed explanations of each component.

\subsubsection{State Encoder} 
The dynamic nature of the driving environment poses a challenge in encoding key driving features, particularly vehicle interactions that indicate safety severity, as both the number and behavior of surrounding obstacles can vary significantly.  
To address this, we draw inspiration from recent advances in object-centric input representations~\cite{renz2023plant}. Following this approach, we adopt an object-centric representation that employs a \textit{Scene Tokenizer} to encode the driving scene as a set of obstacle-level inputs and leverage a \textit{Transformer Encoder}~\cite{vaswani2017attention} to model their interactions. This design also offers greater flexibility in handling highly dynamic driving environments.







\textit{Scene Tokenizer.} 
Given a scene observation $\mathbf{o}_{t}$, the tokenizer transforms the scenario information into a set of object-level representations, denoted as $\hat{\mathbf{o}}_{t} = \{\mathbf{P}_{t} \cup \mathbf{A}_{t}\}$, where $\mathbf{P}_{t} \in \mathbb{R}^{|\mathbb{P}_{t}| \times c}$ and $\mathbf{A}_{t} \in \mathbb{R}^{|\mathbb{A}_{t}| \times c}$ represent the sets of surrounding participants and planned trajectory points at time step $t$, respectively, with each element described by $c = 6$ attributes. 
Specifically, the $i$-th object representation $o_{t}^{i} \in \hat{\mathbf{o}}_{t}$ is defined as:
\[
o_{t}^{i} = \{z_{t}^{i}, x_{t}^{i}, y_{t}^{i}, \varphi_{t}^{i}, w_{t}^{i}, h_{t}^{i}\},
\]
where $z_{t}^{i}$ denotes the category ID of the object (i.e., $z_{t}^{i} = 1$ for surrounding participants and $z_{t}^{i} = 2$ for trajectory points). The remaining attributes describe the object's state: $(x_{t}^{i}, y_{t}^{i})$ represent the position coordinates, $\varphi_{t}^{i}$ denotes the heading angle, and $w_{t}^{i}$ and $h_{t}^{i}$ specify the width and height of the object, respectively.


\textit{Transformer Encoder.} The unified object-level representation $\hat{\mathbf{o}}_{t} \in \mathbb{R}^{(|\mathbb{P}_{t}| + |\mathbb{A}_{t}|) \times 6}$ is then processed by a Transformer encoder, where we adopt a simple BERT architecture~\cite{devlin2019bert}, to extract high-level interaction features.
Following prior practices~\cite{devlin2019bert, dosovitskiy2020image, renz2023plant}, we first project each input token in $\hat{\mathbf{o}}_{t}$ into a hidden embedding space via a linear projection $\rho:\mathbb{R}^{6} \rightarrow \mathbb{R}^{H}$, where $H$ denotes the hidden dimensionality of the Transformer. This results in an embedded sequence of tokens $\mathbf{e}_{t} \in \mathbb{R}^{(|\mathbb{P}_{t}| + |\mathbb{A}_{t}|) \times H}$.
In addition, we use a learnable \texttt{[CLS]} token $e_{\texttt{CLS}} \in \mathbb{R}^{H}$, which is prepended to the sequence of embedded tokens $\mathbf{e}_{t}$ to form the final input sequence $\hat{\mathbf{e}}_{t} \in \mathbb{R}^{(|\mathbb{P}_{t}| + |\mathbb{A}_{t}| + 1) \times H}$. 
The complete sequence $\hat{\mathbf{e}}_{t}$ is then processed by the Transformer encoder, which consists of multiple self-attention layers that model interactions among all tokens while maintaining the same hidden dimensionality $H$. 
After passing through the Transformer layers, the final hidden state corresponding to the \texttt{[CLS]} token is extracted as the global representation of the driving environment, denoted as $\mathbf{x}_{c} \in \mathbb{R}^{H}$. This global feature $\mathbf{x}_{c}$ captures the contextual information of all objects in the scene and is subsequently used to support downstream tasks such as safety assessment and decision repair.

\subsubsection{\monitor}
Safety should be assessed jointly based on the driving environment and the ADS's decision. For example, in a hazardous situation, if the ADS has already executed an emergency braking maneuver, the scenario is still considered safe, as the ADS has taken the correct action to mitigate the risk.  
In contrast, the same situation without an appropriate reaction would be considered dangerous.
Therefore, \monitor{} is designed to predict a safety-critical score for the encoded driving state, incorporating both high-level environmental features extracted by \encoder{} and the corresponding ADS decision. This joint consideration ensures that safety evaluation reflects not only the surrounding environment but also the ADS's actual behavior in response to it.
Formally, given the encoded global representation of the driving environment $\mathbf{x}_{c}$ and the ADS decision $\mathbf{a}_{t}$, the safety-critical score is predicted as:  
\begin{equation}
    \hat{y}_{t}^{\text{safe}} = \text{Sigmoid}(\mathcal{M}_{\phi}(\mathbf{x}_{c} \oplus \mathbf{a}_{t})),
\end{equation}
where $\oplus$ denotes the concatenation operation, $\mathcal{M}_{\phi}(\cdot)$ represents a multi-layer feedforward network parameterized by $\phi$ that maps the concatenated features to a scalar output, and $\text{Sigmoid}(\cdot)$ normalizes the output to the range $(0, 1)$, indicating the severity of the current state's safety. 
A lower value of $y_{\text{safe}}$ corresponds to a safer driving state, while a higher value suggests increased safety risk.

\subsubsection{\repair}
The \repair{} module is designed to predict appropriate actions for mitigating safety risks. In this work, we simplify repair actions to throttle-braking operations for adjusting the longitudinal motion of ADSs~\cite{grieser2020assuring, sun2024redriver, carla_behavior}, focusing on reducing collision risks.  
Specifically, the action space is discretized as $\mathcal{A} = \{0.8, 0.6, \dots, -1.0\}$, dividing the continuous throttle-braking range \([0, 1]\) in the simulator (i.e., CARLA) into ten bins with intervals of $0.2$. Each value in $\mathcal{A}$ represents a normalized throttle-braking intensity, where values less than 0.0 correspond to braking, while others indicate throttle levels.
Formally, given the encoded global representation of the driving environment $\mathbf{x}_{c}$, the repair action is selected as:
\begin{equation}
    \hat{y}_{t}^{\text{repair}} = \arg\max \mathcal{Q}_{\theta}(\mathbf{x}_{c}),
\end{equation}
where $\mathcal{Q}_{\theta}(\cdot)$ is a feedforward network parameterized by $\theta$ that outputs a probability distribution over the action set $\mathcal{A}$, and $\hat{y}_{t}^{\text{repair}}$ denotes the action index corresponding to the highest probability.

\subsubsection{Decision Merger} 
This component merges the ADS decision with the repaired decision predicted by \tool{}. Specifically, the merging process replaces the original ADS decision \( a_{t}^{\text{ads}} \) with the predicted repair decision \( \hat{a}_{t} = \mathcal{A}[\hat{y}_{t}^{\text{repair}}] \) if:
\[ (\hat{y}_{t}^{\text{safe}} > \lambda_{\text{safe}}) \wedge (a_{t}^{\text{ads}} > \hat{a}_{t}) = True
\] The first condition, \( \hat{y}_{t}^{\text{safe}} > \lambda_{\text{safe}} \), indicates that the driving state requires intervention if the predicted safety-critical score \(\hat{y}_{t}^{\text{safe}}\) exceeds a predefined threshold \(\lambda_{\text{safe}}\). The second condition, \( a_{t}^{\text{ads}} > \hat{a}_{t} \), ensures that the ADS decision is replaced only if it is less conservative than \tool{}'s repair decision in safety-critical situations, thereby preserving the original ADS decision when it has already taken sufficient precautionary maneuvers.

\begin{figure}[!t]
    \centering
    \includegraphics[width=1.0\linewidth]{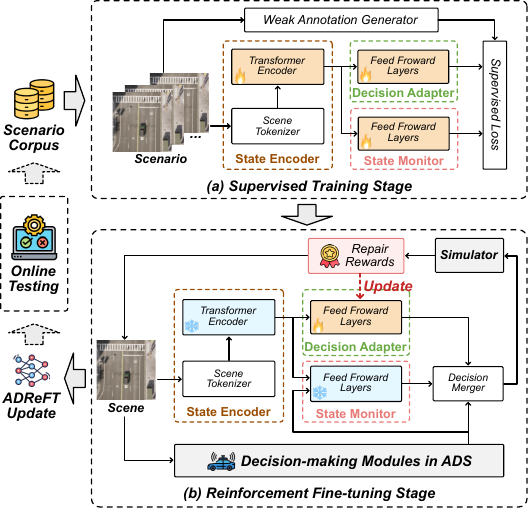}
    \caption{Illustration of training \tool.}
    \label{fig:training}
\end{figure}
\subsection{Training of \tool}
Figure~\ref{fig:training} illustrates the overall training pipeline for \tool, which utilizes known testing scenarios.  
Specifically, as shown in Figure~\ref{fig:training}(a), training begins with a warm-up \textit{Supervised Training Stage}, where all parameters of \tool{} are optimized using weak supervision through supervised learning. 
This stage equips \tool{} with a fundamental ability to identify safety-critical states while ensuring safety through conservative repair decisions, i.e., predicting the most intensive repair action.  
Since overly intensive repair behaviors can excessively impact ADS motion, we introduce the \textit{Reinforcement Fine-tuning Stage} (ReFT) to mitigate these limitations and enable more adaptive, context-aware repair strategies.  
As illustrated in Figure~\ref{fig:training}(b), this stage updates only the trainable parameters within the \repair{} component while freezing all other parameters, allowing it to explore adaptive repair actions without degrading the fundamental ability to identify safety-critical states.  
The ReFT process is guided by a \textit{safe-explore} step reward and a termination reward derived from execution results (e.g., successful repairs).  
Finally, through this two-stage training process, \tool{} continuously adapts to improve online safety while minimizing intervention intensity in the ADS's original decision-making process.  

\subsubsection{Supervised Training Stage.}
One of the most significant challenges in this stage is obtaining fine-grained annotations for each scene in scenarios. This difficulty arises because a scenario typically consists of thousands of frames, with only a final outcome indicating the testing results, such as success or violation.  
To address this limitation, we design a \textit{Weak Annotation Generator} to automatically produce weak annotations for each frame based on the scenario outcome and predefined regulations. Although these annotations are not perfectly accurate, they are designed to guide the repair of potential safety-critical states as comprehensively as possible.  
Subsequently, we use these weak annotations to train \tool{}, allowing it to learn to identify safety-critical states and predict corresponding repair actions.  

\textit{Weak Annotation Generator.} Given a scenario \( S \) consisting of \( T \) individual scenes, we define the weak label assignment function for safety severity as \( \mathcal{F}_{\text{safe}}: \{1, \dots, T\} \to \{0,1\} \), formulated as:
\begin{equation}\label{eq:annotation}
    y_t^{\text{safe}} = \mathcal{F}_{\text{safe}}(t) =
    \begin{cases}
        1, & \text{if } d_t < \delta_{d} \vee (Y_S = 1 \wedge t > T - \delta_{t}), \\
        0, & \text{otherwise}.
    \end{cases}
\end{equation}
Here, \( t \) denotes the timestamp of the scene, and \( Y_S \) represents the scenario result, where \( Y_S = 1 \) indicates that a collision has occurred, while \( Y_S = 0 \) denotes a successful scenario without violations. 
The variable \( d_t \) represents the minimum distance between the ego vehicle and surrounding obstacles at timestamp \( t \). 
\( \delta_d \) and \( \delta_t \) are hyperparameters that define the labeling criteria. 
Therefore, a scene is considered safety-critical ($y_{t}^{\text{safe}}=1$) if the minimum distance falls below the safety threshold \( \delta_d \) or if it occurs within \( \delta_t \) timesteps before an inevitable collision (\( Y_S = 1 \)). In such situations, additional safety mechanisms should be employed to ensure safe operation.  
Note that a scenario terminates at \( T \) either when the ego vehicle collides with an obstacle or successfully reaches its destination.

Subsequently, we assign a default conservative repair action annotation to each scene as follows:  
\begin{equation}
    y_{t}^{\text{repair}} =
    \begin{cases}
        \operatorname{argmax}(\mathcal{A}), & \text{if } y_{t}^{\text{safe}} = 1, \\
        \operatorname{argmin}(\mathcal{A}), & \text{otherwise}.
    \end{cases}
\end{equation}
where \( \mathcal{A} \) represents the discrete repair action space. The maximum index in \( \mathcal{A} \) corresponds to the strongest repair action (e.g., maximum braking), while the minimum index represents a more relaxed repair action (e.g., slight throttle release).  
Here, we assign the strongest repair action to scenes labeled as safety-critical (\( y_{t}^{\text{safe}} = 1 \)), ensuring maximal intervention in high-risk situations. Conversely, for non-safety-critical scenes, no repair action or a relaxed repair action is assigned, indicating that intervention is unnecessary.  

\textit{Training Objective.} The training objective for optimizing the parameters of \tool{} is to minimize the discrepancy between the model predictions and the assigned weak annotations. 
Specifically, this objective is defined over the pairs (\(\hat{y}^{\text{safe}}, {y}^{\text{safe}}\)) and (\(\hat{y}^{\text{repair}}, y^{\text{repair}}\)), where the first pair corresponds to safety-critical state classification and the second to repair action classification.  
The loss function is formulated as:
\begin{equation}
    L_{\text{SL}} = \frac{1}{N} \sum_{i=1}^{N} \Big[ L_{\text{BCE}}(y_{i}^{\text{safe}}, \hat{y}_{i}^{\text{safe}}) + L_{\text{CE}}(y_{i}^{\text{repair}}, \hat{y}_{i}^{\text{repair}}) \Big],
\end{equation}
where \( L_{\text{BCE}} \) denotes the binary cross-entropy loss for safety-critical state classification, and \( L_{\text{CE}} \) represents the cross-entropy loss for repair action classification. The total loss \( L_{\text{SL}} \) is computed as the average loss over \( N \) training samples.

\begin{algorithm}[!t]
\small
\SetKwInOut{Input}{Input}
\SetKwInOut{Output}{Output}
\SetKwInOut{Para}{Parameters}
\SetKwComment{Comment}{\color{blue}// }{}
\SetKwFunction{Train}{{\textbf{Training}}}
\SetKwFunction{RF}{\textbf{RewardCalculator}}
\SetKwProg{Fn}{Function}{:}{}
\Input{
Scenario corpus $\mathbf{S}_{\text{corpus}}$ \\ 
Warm-started \tool $\mathcal{F}_{\tool}$
}
\Output{
Fine-tuned \tool{} $\mathcal{F}_{\tool}$
}

\Fn{\Train{}}{
    Initialize an empty replay buffer $\mathcal{D} \gets \{\}$ with max length $N_{\mathcal{D}}$ \\
    
    \For{episode step e = 1, 2, \ldots, E}{
        $t = 0$ \Comment{reset the step counter}
        $S_{e} \gets \text{Sample}(\mathbf{S}_{\text{corpus}})$ \\
        $o_{t} \gets \text{Simulator}.{reset}(S_{e})$ \Comment{reset simulator}
        \Repeat{Scenario $S_{e}$ terminates}{ 
            $\{\hat{a}_{t}, a_{t}^{\text{ads}}, \hat{y}_{t}^{\text{safe}},  \hat{y}_{t}^{\text{repair}}\} \gets \mathcal{F}_{\tool}({o}_{t})$ \\
            $o_{t+1}, \mathbf{z} \gets \text{Simulator}.{step}(\hat{a}_{t})$ \Comment{tick simulator}
            
            $r_t \gets $ \RF{$o_{t+1}, \hat{a}_{t}, a_{t}^{\text{ads}}, \hat{y}_{t}^{\text{safe}}, \mathbf{z}$} \\
            $\mathcal{D} \gets \mathcal{D} \cup \{o_{t}, \hat{y}_{t}^{\text{repair}}, r_{t}, o_{t+1}\}$ \\
            $t = t + 1$ \Comment{count the simulation steps}
        }
        Only update $\mathcal{Q}_{\theta}$ in $\mathcal{F}_{\tool}$ by Eq.~\eqref{eq:DQN_Q} and Eq.~\ref{eq:DQN_Loss} 
    }
    \Return $\mathcal{F}_{\tool}$
}

\Fn{\RF{$o_{t+1}, \hat{a}_{t}, a_{t}^{\text{ads}}, \hat{y}_{t}^{\text{safe}},  \mathbf{z}$}}{
    $ r_{t} = (1-\hat{y}_{t}^{\text{safe}}) \times (1 - 0.5 |a_{t}^{\text{ads}} - \hat{a}_{t}|)$ \\
        

    \If{$\mathbf{z} \  \text{contains violations}$}{
        \Comment{Calculate termination rewards}
        $r_t = r_t - 10.0$ \\
    }
    \Return $r_t$ 
} 
\caption{Reinforcement Fine-tuning for \tool}
\label{algo:reft}
\end{algorithm}
\subsubsection{Reinforcement Fine-tuning.}
Supervised training with weak annotations makes \tool{} overly conservative, often leading to suboptimal decisions that consistently favor extreme repair actions, such as applying maximum braking in all identified safety-critical states. 
This overly intensive behavior can excessively disrupt the original ADS driving experience, potentially resulting in frequent uncomfortable motions. 
However, a key challenge lies in identifying more suitable and adaptive repair decisions that ensure safety while minimizing intervention intensity to the original ADS's motion. 
To address this, we introduce \textit{Reinforcement Fine-Tuning (ReFT)}, inspired by reinforcement learning, which fine-tunes only \repair{} to explore adaptive repair strategies based on feedback from replayed experiences.  

Algorithm~\ref{algo:reft} illustrates the \textit{ReFT} process, which iteratively replays cases in the scenario corpus $\mathbf{S}_{\text{corpus}}$ using an ADS equipped with the warm-started \tool{} model $\mathcal{F}_{\tool}$, trained in the previous stage. Note that the scenario corpus saves testing cases collected from ADS testing techniques. 
The model is fine-tuned with designed rewards until the maximum number of episodes $E$ is reached (Lines 3-14).
In each episode $e$, a scenario $S_{e}$ is first sampled from $\mathbf{S}_{\text{corpus}}$ (Line 5), and the simulation environment is reset (Line 6). 
During each execution step $t$, the execution states ($o_{t}$, $o_{t+1}$), the predicted repair decision $\hat{y}_{t}^{\text{repair}}$ made by \tool{}, and the step reward $r_{t}$ are recorded in the replay buffer $\mathcal{D}$ (Lines 7-13) until the scenario terminates, either by reaching the destination or encountering a violation. 
Once the scenario $S_{e}$ terminates (Line 13), only the parameters in \repair{} are updated based on the DQN algorithm, as detailed in Eq.~\eqref{eq:DQN_Q} and Eq.~\eqref{eq:DQN_Loss} in Section~\ref{sec:background_RL}. The ReFT process finally concludes by outputting the fine-tuned \tool{} (Line 15).

\textit{Reward Function.} 
The reward function is a crucial component in reinforcement learning. At this stage, as shown in Lines 16–20 of Algorithm~\ref{algo:reft}, we design a reward function that accounts for two key aspects: \textit{safe-explore reward} and \textit{termination reward}.
(1) The \textit{safe-explore reward} (Line 17) is designed to encourage \tool{} to explore repair actions that are less intensive in safer driving states. Specifically, we measure the intensity by calculating the L1-distance between the original ADS decision $a_{t}^\text{ads}$ and the repaired decision $\hat{a}_{t}$ made by \tool{}, denoted as $0.5|a_{t}^\text{ads}-\hat{a}_{t}|$. We normalize this value to the range [0, 1] by multiplying by 0.5. Therefore, the explore reward is calculated by $1-0.5|a_{t}^\text{ads}-\hat{a}_{t}|$, where a larger value indicates less intensive repairs. 
To prioritize safety, we use the pretrained \monitor{} as a reward function to estimate the surrounding safety, which serves as a weighting factor $1-\hat{y}_{t}^{\text{safe}}$ to scale the explore reward, making safer situations yield a larger reward and more dangerous situations yield a smaller explore reward. Therefore, a higher reward is assigned when the surroundings are safe, while ensuring minimal intensive repair actions from the original ADS decision.
(2) The \textit{termination reward} (Line 18-19) assesses the effectiveness of the repair based on the execution outcome. If the ADS equipped with \tool{} results in a violation, the repair is considered ineffective, and the agent receives a fixed high penalty (e.g., -10.0). 
Ultimately, these two rewards are summed to guide the training process.

\begin{table*}[!t]
    \centering
    \caption{Comparison of online repair effectiveness with baselines.}
    \vspace{-10pt}
    \small
    \resizebox{1.0\linewidth}{!}{
        \begin{tabular}{l|l|ccc|ccc|ccc|ccc|ccc|ccc}
        \toprule
         \multirow{2.5}*{\textit{ADS}} & \multirow{2.5}*{\textit{Method}} &\multicolumn{3}{c|}{\textit{S1}} &\multicolumn{3}{c|}{\textit{S2}} &\multicolumn{3}{c|}{\textit{S3}} &\multicolumn{3}{c|}{\textit{S4}} &\multicolumn{3}{c|}{\textit{S5}} &\multicolumn{3}{c}{\textit{Average}}\\
         \cmidrule(lr){3-5}\cmidrule(lr){6-8}\cmidrule(lr){9-11}\cmidrule(lr){12-14}\cmidrule(lr){15-17}\cmidrule(lr){18-20}
         &  & {\textit{\%Fix}$\uparrow$} & {\textit{\%Deg.}$\downarrow$} &  \cellcolor{lightgray!30}\textit{$\Delta E\uparrow$} & {\textit{\%Fix}$\uparrow$} & {\textit{\%Deg.}$\downarrow$} &  \cellcolor{lightgray!30}\textit{$\Delta E\uparrow$} &{\textit{\%Fix}$\uparrow$} & {\textit{\%Deg.}$\downarrow$} &  \cellcolor{lightgray!30}\textit{$\Delta E\uparrow$} &{\textit{\%Fix}$\uparrow$} & {\textit{\%Deg.}$\downarrow$} &  \cellcolor{lightgray!30}\textit{$\Delta E\uparrow$} &{\textit{\%Fix}$\uparrow$} & {\textit{\%Deg.}$\downarrow$} &  \cellcolor{lightgray!30}\textit{$\Delta E\uparrow$} &{\textit{\%Fix}$\uparrow$} & {\textit{\%Deg.}$\downarrow$} &  \cellcolor{lightgray!30}\textit{$\Delta E\uparrow$}  \\
        \midrule

        \multirow{4}*{\textit{Roach}} & \textit{Random} & 45.0 & 7.5 & \cellcolor{lightgray!30}37.5 & 35.0 & 5.0 & \cellcolor{lightgray!30}30.0 & 40.0 & 5.0 & \cellcolor{lightgray!30}35.0 & 7.5 & 12.5 & \cellcolor{lightgray!30}-5.0 & 7.5 & 2.5 & \cellcolor{lightgray!30}5.0 & 27.0 & 6.5 & \cellcolor{lightgray!30}20.5 \\ 
        
        & \textit{RuleRepair} & 57.5 & \textbf{0.0} & \cellcolor{lightgray!30}57.5 & 82.5 & \textbf{0.0} & \cellcolor{lightgray!30}82.5 & 47.5 & 5.0 & \cellcolor{lightgray!30}52.5 & 62.5 & 10.0 & \cellcolor{lightgray!30}52.5 & 40.0 & \textbf{0.0} & \cellcolor{lightgray!30}40.0 & 58.0 & \textbf{3.0} & \cellcolor{lightgray!30}55.0 \\
        
        & \textit{ModelRepair} & 72.5 & 5.0 & \cellcolor{lightgray!30}67.5 & 70.0 & 15.0 & \cellcolor{lightgray!30}55.0 & 85.0 & \textbf{2.5} & \cellcolor{lightgray!30}82.5 & 27.5 & 12.5 & \cellcolor{lightgray!30}15.0 & 27.5 & 2.5 & \cellcolor{lightgray!30}25.0 & 56.5 & 7.5 & \cellcolor{lightgray!30}49.0 \\
        
        & \textit{\tool} & \textbf{82.5} & 12.5 & \cellcolor{lightgray!30}\textbf{70.0} & \textbf{97.5} & 5.0 & \cellcolor{lightgray!30}\textbf{92.5} & \textbf{90.0} & \textbf{2.5} & \cellcolor{lightgray!30}\textbf{87.5} & \textbf{85.0} & \textbf{7.5} & \cellcolor{lightgray!30}\textbf{77.5} & \textbf{97.5} & \textbf{0.0} & \cellcolor{lightgray!30}\textbf{97.5} & \textbf{90.5} & 5.5 & \cellcolor{lightgray!30}\textbf{85.0} \\
        \midrule

         \multirow{4}*{\textit{Pylot}} & \textit{Random} & 37.5 & \textbf{0.0} & \cellcolor{lightgray!30}37.5 & 12.5 & 2.5 & \cellcolor{lightgray!30}10.0 & 15.0 & \textbf{0.0} & \cellcolor{lightgray!30}15.0 & 5.0 & 7.5 & \cellcolor{lightgray!30}-2.5 & 12.5 & 2.5 & \cellcolor{lightgray!30}10.0 & 16.5 & 2.5 & \cellcolor{lightgray!30}14.0 \\ 
         
        & \textit{RuleRepair} & 5.0 & 0.0 & \cellcolor{lightgray!30}5.0 & 65.0 & \textbf{0.0} & \cellcolor{lightgray!30}65.0 & 50.0 & \textbf{0.0} & \cellcolor{lightgray!30}50.0 & 15.0 & 7.5 & \cellcolor{lightgray!30}7.5 & 75.0 & \textbf{0.0} & \cellcolor{lightgray!30}75.0 & 42.0 & \textbf{1.5} & \cellcolor{lightgray!30}40.5 \\
        
        & \textit{ModelRepair} & 60.0 & 5.0 & \cellcolor{lightgray!30}55.0 & 42.5 & 2.5 & \cellcolor{lightgray!30}40.0 & 42.5 & 2.5 & \cellcolor{lightgray!30}40.0 & 20.0 & 10.0 & \cellcolor{lightgray!30}10.0 & 7.5 & 2.5 & \cellcolor{lightgray!30}5.0 & 34.5 & 4.5 & \cellcolor{lightgray!30}30.0 \\
        
        & \textit{\tool} & \textbf{72.5} & 2.5 & \cellcolor{lightgray!30}\textbf{70.0} & \textbf{85.0} & 0.0 & \cellcolor{lightgray!30}\textbf{85.0} & \textbf{60.0} & 5.0 & \cellcolor{lightgray!30}\textbf{55.0} & \textbf{80.0} & \textbf{5.0} & \cellcolor{lightgray!30}\textbf{75.0} & \textbf{95.0} & \textbf{0.0} & \cellcolor{lightgray!30}\textbf{95.0} & \textbf{78.5} & 2.5 & \cellcolor{lightgray!30}\textbf{76.0} \\

        \bottomrule
        \end{tabular}
    }
    \vspace{-5pt}
    \label{tab:rq1-repair}
\end{table*}

\section{Evaluation}

In this section, we aim to empirically evaluate the capability of \tool in repairing violations made by ADSs. In particular, we will answer the following research questions:

\noindent \textbf{RQ1}: Can \tool{} effectively repair safety-critical violations while minimizing its impact on the original ADS?

\noindent \textbf{RQ2}: How effective are the components and training stages of \tool{} in improving repair performance?

\noindent \textbf{RQ3}: How does \tool{} perform in terms of computational cost?

To answer these research questions, we conduct experiments using the following settings:

\textbf{Environment.} We integrated \tool within the CARLA~\cite{dosovitskiy2017carla} simulator using its Python API, aiming to assess its applicability across diverse ADS architectures. Our evaluation encompasses two ADSs, Roach~\cite{roach_iccv} and Pylot~\cite{gog2021pylot}, each of which has been extensively explored in prior research~\cite{xu2022safebench, icse_samota, haq2023many}. Roach is an end-to-end ADS that demonstrates expert-level driving by processing perfect perception data to make driving decisions. Pylot is a module-based ADSs including localization, perception, prediction, planning and control modules, which demonstrates stable autonomous driving performance.
To concentrate on decision repair, we ensure all ADSs are provided with perfect perception input from the simulator.

\textbf{Benchmark.} To the best of our knowledge, no publicly available repair benchmark exists for ADS testing, and violations depend on the ADS under test.  
Thus, we use existing testing techniques, such as \textit{Random} and \textit{BehAVExplor}~\cite{cheng2023behavexplor}, to gather violations and normal test cases from five representative scenarios in the NHTSA pre-crash typology~\cite{najm2007pre}, including left turn (\textit{S1}), right turn (\textit{S2}), crossing intersection (\textit{S3}), highway exit (\textit{S4}), and on-ramp merging (\textit{S5}), which are widely adopted in previous studies~\cite{av_fuzzer, cheng2023behavexplor, thorn2018framework, zhou2023specification, icse_samota, huai2023doppelganger, huai2023sceno, css_drivefuzzer}.
In total, for each ADS, we selected 400 verification scenarios, each containing 40 violations and 40 success cases to assess repair performance.  
Notably, the synchronization mechanisms in CARLA, Roach, and Pylot mitigate simulator non-determinism, ensuring that discovered violations can be reliably reproduced through re-simulation.

\textbf{Baselines.} In this paper, we select two types of baselines: model-based and rule-based approaches.  
To the best of our knowledge, existing model-based approaches~\cite{stocco2020misbehaviour, stocco2022thirdeye} focus solely on safety monitors for L2 Advanced Driver Assistance Systems (ADASs), which cannot be directly used in our work.
Thus, we adopt their core algorithms of detecting out-of-distribution samples and integrate an emergency braking operation as our model-based baseline, denoted as \textit{ModelRepair}.  
Previous rule-based online repair techniques~\cite{carla_behavior, grieser2020assuring, sun2024redriver} rely on predefined specifications (i.e., traffic rules) that are not directly compatible with the CARLA platform and the selected ADSs for repairing safety-critical violations, particularly collisions targeted in this study.  
To address this, following their implementations, we develop a rule-based repair baseline, denoted as \textit{RuleRepair}, using the widely adopted time-to-collision (TTC)~\cite{carla_behavior, grieser2020assuring} as the safety-critical specification and integrating it with the same repair action space as \tool{} for consistency in the comparison.  
Moreover, we include a \textit{Random} baseline that selects driving scenes and applies a random repair decision from the same repair action space as \tool{} for reference.  


\textbf{Metrics.} \label{metric}
In our experiments, we adopt three primary metrics to assess repair effectiveness: \textit{\%Fixed}, \textit{\%Degraded}, and $\Delta E$.  
Specifically, \textit{\%Fixed} quantifies the percentage of violations in the testing set that are successfully repaired, directly reflecting the repair capability.  
\textit{\%Degraded} measures the percentage of previously successful scenarios that introduce new violations after repair, highlighting potential negative impacts.  
$\Delta E$ represents the overall improvement brought by the repair method to the ADS, with a larger value indicating greater effectiveness.
Additionally, we assess the impact of repair actions by introducing \textit{\%Intensity}, a metric that quantifies the deviation between the repair decision and the original ADS decision over the course of the ADS motion. This metric provides insight into the extent of intervention, where a lower value indicates that the repair decision minimally perturbs the original driving behavior, thereby preserving the natural driving experience of the ADS.

\textbf{Training Details.} The \textit{Supervised Training Stage} runs for 200 epochs with a learning rate of \(10^{-4}\). During \textit{Reinforcement Fine-tuning}, only the parameters in \repair{} remain trainable. We employ \(\varepsilon\)-Greedy exploration, where actions are sampled uniformly at random from the action space. The training process consists of 300 episodes, utilizing a replay memory \( \mathcal{D} \) of size 5,000, and a learning rate of \(10^{-4}\). In each episode, we randomly sample both success and violation scenarios from the corpus for replay.

\textbf{Implementation.} 
Following previous works~\cite{stocco2020misbehaviour, stocco2022thirdeye}, we set the safety-critical threshold \( \lambda_{\text{safe}} \) at the safety score such that more than 95\% of positive samples are correctly recognized.
For weak annotation parameters, we empirically set \( \delta_{d} \) to 1 meter and \( \delta_{t} \) to 3 seconds, based on~\cite{lee2013study, zhao2017review}.  

\subsection{RQ1: Effectiveness of \tool}

\subsubsection{Repair Effectiveness} Table~\ref{tab:rq1-repair} presents a comparative results of online repair effectiveness against baseline methods across the testing set, covering five scenarios (\textit{S1} to \textit{S5}).  
For both the end-to-end ADS (\textit{Roach}) and the module-based ADS (\textit{Pylot}), \tool achieves the best repair performance, improving ADS safety by an average of 85\% for \textit{Roach} and 76\% for \textit{Pylot}. This significantly outperforms the best baseline, \textit{RuleRepair}, which achieves 55.0\% for \textit{Roach} and 40.5\% for \textit{Pylot}.  
By analyzing the results of \textit{Random}, we find that randomly perturbing the ADS decision may fix some violations. However, the uncontrollable uncertainty makes it impractical for real-world usage. For example, it degrades the overall performance of both ADSs in \textit{S4}, with a reduction of 5.0\% for \textit{Roach} and 2.5\% for \textit{Pylot}.
Comparing \tool with \textit{RuleRepair}, we observe that \tool consistently outperforms this baseline in all scenarios, particularly in fixing a greater number of violations, as indicated by the columns under \textit{\%Fix}.
However, we also find that \textit{RuleRepair} maintains a relatively low \textit{\%Degraded} compared to \tool and other baselines.  
This suggests that \textit{RuleRepair} is limited to repairing violations that match predefined rule-based patterns, resulting in a lower recall of safety-critical states and, consequently, fewer interventions in various scenarios.  
Nevertheless, its lower \textit{\%Fix} and $\Delta E$, particularly in \textit{S1} (5\%) and \textit{S4} (7.5\%) for \textit{Pylot}, indicate limited generalization and effectiveness in repairing ADSs across different scenarios.
We also find that \textit{ModelRepair} performs worse than all other baselines except \textit{Random}, particularly with performance lower than the rule-based repair method (\textit{Roach}: 49.0\% vs. 55.0\%, \textit{Pylot}: 30.0\% vs. 40.5\%).  
This is because the adopted model~\cite{stocco2020misbehaviour} treats out-of-distribution states as safety-critical, which is not well-suited for repairing L4 autopilot safety-critical issues caused by decision-making modules. These issues are primarily triggered by small perturbations, leading abnormal states to exhibit high distributional similarity to normal states.
Overall, \tool demonstrates the highest effectiveness in repairing safety-critical violations identified during ADS testing, indicating a powerful method for automatically maintaining safety.

\subsubsection{Repair Intensity.} We further evaluate the intervention intensity (\textit{\%Intensity}) of \tool and baseline methods, which quantifies the deviation between the throttle-brake commands originally decided by the ADS and those modified by \tool. This metric ranges from [0, 2], where a lower value indicates a more effective repair with minimal disruption to the ADS’s original driving behavior.  
Table~\ref{rq1:intensity} presents the detailed results. On average, we find that \tool achieves a lower \textit{\%Intensity}, with 0.77 for \textit{Roach} and 0.72 for \textit{Pylot}, outperforming the best baseline—\textit{Random} for \textit{Roach} (0.78) and \textit{RuleRepair} for \textit{Pylot} (0.91). This indicates that \tool introduces fewer interventions in original ADS decisions, thereby minimizing disruptions to vehicle motion.
Furthermore, we observe that \textit{Random} exhibits a relatively lower \textit{\%Intensity} on \textit{Roach} compared to \tool in \textit{S1} (0.83 vs. 0.87), \textit{S2} (0.83 vs. 0.93), and \textit{S3} (0.81 vs. 0.88). Additionally, \textit{RuleRepair} achieves a lower \textit{\%Intensity} of 0.39 in \textit{S1} compared to \tool’s 0.71 for \textit{Pylot}. 
However, considering the lower repair effectiveness of \textit{Random} and \textit{RuleRepair} reported in Table~\ref{tab:rq1-repair}, we conclude that minimizing intervention intensity alone does not necessarily lead to better repair performance. Instead, this highlights that \tool can generate adaptive repair decisions based on the driving context, effectively balancing intervention and safety improvements. 

\begin{table}[!t]
    \centering
     \caption{Comparison of \textit{\%Intensity} with baselines}
     \vspace{-10pt}
     \resizebox{0.9\linewidth}{!}{
    \begin{tabular}{l|l|ccccc|c}
    \toprule
    \textit{ADS} & \textit{Method} & \textit{S1} & \textit{S2} & \textit{S3} & \textit{S4} & \textit{S5} & \cellcolor{lightgray!30}\textit{Avg.}  \\
    \midrule
    
    \multirow{4}*{\textit{Roach}} & \textit{Random} & \textbf{0.83} & \textbf{0.83} & \textbf{0.81} & 0.69 & 0.73 & \cellcolor{lightgray!30}0.78 \\

    & \textit{RuleRepair} & 0.91 & 1.39 & 1.11 & 1.35 & 0.79 & \cellcolor{lightgray!30}1.11 \\
    
    & \textit{ModelRepair} & 1.75 & 1.79 & 1.65 & 1.55 & 1.63 & \cellcolor{lightgray!30}1.67 \\
    
    & \tool & 0.87 & 0.93 & 0.88 & \textbf{0.61} & \textbf{0.56} & \cellcolor{lightgray!30}\textbf{0.77} \\
    \midrule

    \multirow{4}*{\textit{Pylot}} & \textit{Random} & 1.23 & 1.23 & 1.18 & 1.16 & 1.14 & \cellcolor{lightgray!30}1.20 \\
    
    & \textit{RuleRepair} & \textbf{0.39} & 1.21 & 0.72 & 1.32 & 0.89 & \cellcolor{lightgray!30}0.91 \\
    
    & \textit{ModelRepair} & 1.77 & 1.71 & 1.71 & 1.65 & 1.61 & \cellcolor{lightgray!30}1.69 \\
    
    & \tool & 0.71 & \textbf{0.74} & \textbf{0.63} & \textbf{0.78} & \textbf{0.74} & \cellcolor{lightgray!30}\textbf{0.72} \\
    \bottomrule
    \end{tabular}
    }
    \label{rq1:intensity}
\end{table}
\begin{figure*}[!t]
    \centering
    \includegraphics[width=1.0\textwidth]{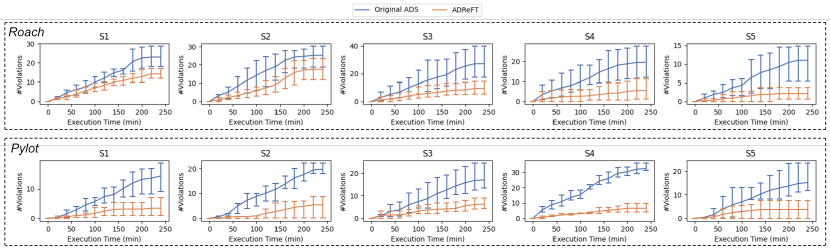}
    \vspace{-20pt}
    \caption{\textit{Robustness Testing for \tool{}.} The X-axis represents the execution time of ADS testing, while the Y-axis indicates the number of detected violations.}
    \label{fig:robust}
\end{figure*}

\subsubsection{Robustness Testing} 
To assess the robustness of \tool, we conducted online testing using existing ADS testing techniques. Specifically, we selected three widely used ADS testing techniques: \textit{Random}, \textit{AVFuzzer}~\cite{av_fuzzer}, and \textit{BehAVExplor}~\cite{cheng2023behavexplor}.  
We ran each technique three times across the five scenarios (\textit{S1} to \textit{S5}), evaluating both the \textit{Original ADS} without any repair mechanism and the ADS equipped with \tool.  
Figure~\ref{fig:robust} illustrates the mean and min-max variance of detected safety-critical violations by these three testing techniques over time (in minutes).  
For both \textit{Roach} and \textit{Pylot}, we observed a notable increase in the number of violations in the \textit{Original ADS} as fuzzing time progressed.  
Conversely, the ADS equipped with \tool maintained a stable and lower number of violations throughout the tests, demonstrating its robustness and reliability in enhancing online safety for ADSs.

\begin{center}
\vspace{-10pt}
\fcolorbox{black}{gray!10}{\parbox{\linewidth}{\textbf{Answer to RQ1}: \tool significantly outperforms baseline methods in repairing violations across different ADSs, effectively improving ADS performance by 80.5\% while adaptively making decisions to maintain a lower intervention degree of 0.74. Additionally, \tool demonstrates robustness against existing ADS testing techniques.
}}
\end{center}

\subsection{RQ2: Effectiveness of Components}\label{rq2-ablation}
In this section, we further investigate the effectiveness of each component in \tool, including \monitor and \repair, as well as the design of training strategies.

\begin{table}[!t]
    \centering
    \caption{Effectiveness of State Monitor}
    \vspace{-10pt}
    \resizebox{0.8\linewidth}{!}{
    \begin{tabular}{l|l|ccc}
    \toprule
      \textit{ADS} & \textit{Method} & \textit{Prec.} & \textit{Recall} & \textit{F1}  \\
     \midrule
     \multirow{4}*{\textit{Roach}} & \textit{RuleBase} & 52.78 & 9.15 & 15.60 \\
     & \textit{SelfOracle} & 12.21 & 10.71 & 11.41 \\
     & \textit{wo Regulation} & \textbf{60.01} & 68.68 & 64.05 \\
     & \textit{\tool} & 55.43 & \textbf{80.09} & \textbf{65.52} \\
     \midrule

     \multirow{4}*{\textit{Pylot}} & \textit{RuleBase} & 74.42 & 6.91 & 12.64\\
     & \textit{SelfOracle} & 18.80 & 26.75 & 22.08 \\
     &  \textit{wo Regulation} & 51.30 & 79.96 & 62.50 \\
     
     &  \tool & \textbf{51.90} & \textbf{80.34} & \textbf{63.06} \\
     
     
     
     \bottomrule
    \end{tabular}
    }
    \label{rq2:monitor}
\end{table}

\subsubsection{Effectiveness of State Monitor}\label{sec: Monitor_effect} 
To evaluate the effectiveness of \monitor, we compare its ability to detect safety-critical states against three baseline methods:  
(1) \textit{RuleBase}, which uses predefined regulations~\cite{carla_behavior, grieser2020assuring, sun2024redriver} (e.g., time-to-collision) to identify safety-critical states;  
(2) \textit{SelfOracle}~\cite{stocco2020misbehaviour}, which trains a surrogate model to detect out-of-distribution states. Note that \textit{SelfOracle} was originally designed for L2 ADAS systems with image input; in our implementation, we adapt it by feeding Bird’s-Eye-View images and leverage it for reference comparison;  
(3) \textit{wo Regulation}, which trains \tool without annotations derived from safety regulations (i.e., $d_t < \delta_{d}$) in Eq.~\ref{eq:annotation}.  
Specifically, we label states occurring within 3.0 seconds before a collision as positive samples (i.e., safety-critical states), indicating that the monitor should predict collisions in advance. All other states are annotated as negative samples.
Table~\ref{rq2:monitor} reports the \textit{Precision (Prec.)}, \textit{Recall}, and \textit{F1} score across all cases in the test set.  
According to previous studies~\cite{stocco2020misbehaviour, stocco2022thirdeye}, \textit{Recall} is crucial for evaluating the monitor's ability to identify safety-critical states without missing them. We observe that \tool achieves the highest \textit{Recall} and overall \textit{F1} score for both ADSs, demonstrating its superior performance in detecting safety-critical states.  
Specifically, we find that \textit{RuleBase} captures only a limited number of safety-critical states, achieving a \textit{Recall} of just 9.15 for \textit{Roach} and 6.91 for \textit{Pylot}. This limitation arises because defining explicit rule-based specifications for collisions in dynamic driving environments is inherently challenging.  
Comparing \textit{SelfOracle}, we observe that its lower performance is due to our focus on decision-making-induced failures, which are typically caused by small perturbations that are not easily captured by the Bird’s-Eye-View (BEV) representation fed to \textit{SelfOracle}. These perturbations are difficult to distinguish from normal states using out-of-distribution detection alone, limiting the effectiveness of \textit{SelfOracle}.  
Furthermore, we find that training \tool with additional regulatory annotations improves its ability to accurately identify more safety-critical states. In detail, \textit{wo Regulation} achieves only 68.68 \textit{Recall} for \textit{Roach} and 79.96 for \textit{Pylot}, whereas \tool achieves higher recall scores of 80.09 and 80.34, respectively. Additionally, \tool achieves improved precision (51.90 vs. 51.30 for \textit{Roach}) and comparable precision (55.43 vs. 60.01 for \textit{Pylot}), demonstrating that incorporating some rule-based safety constraints enhances detection performance.

\begin{table}[!t]
    \centering
    \caption{Effectiveness of Repair Strategies}
    \vspace{-10pt}
    \resizebox{0.85\linewidth}{!}{
    \begin{tabular}{l|l|ccc|c}
    \toprule
      \textit{ADS} & \textit{Method} & \textit{\%Fix} & \textit{\%Deg.} & \textit{$\Delta E$} & \textit{\%Intensity}  \\
     \midrule
     
     \multirow{5}*{\textit{Roach}} & \textit{Rand-R} & 80.0 & 16.5 & 63.5 & 0.89 \\
     
     & \textit{only SL} & 87.5 & 12.0 & 75.5 & 0.98 \\
     
     & \textit{only RL} & 36.5 & 6.0 & 30.5 & \textbf{0.19} \\
     
     & \textit{wo SR} & 87.0 & 7.5 & 79.5 & 0.83 \\
     
     & \textit{\tool} & \textbf{90.5} & \textbf{5.5} & \textbf{85.0} & {0.77} \\
     \midrule
     
     \multirow{5}*{\textit{Pylot}} & \textit{Rand-R} & 63.0 & 6.5 & 56.5 & 0.93 \\
     & \textit{only SL} & 72.0 & \textbf{0.0} & 72.0 & 0.87 \\
     
     & \textit{only RL} & 32.0 & 3.5 & 28.5 & 0.23 \\
     
     & \textit{wo SR} & 75.5 & 3.5 & 72.0 & 1.02 \\
     
     & \textit{\tool} & \textbf{78.5} & 2.5 & \textbf{76.0} & 0.72 \\
     \bottomrule
    \end{tabular}
    }
    \label{tab:rq2-repair}
\end{table}

\subsubsection{Effectiveness of Repair Strategies} 
To evaluate the effectiveness of {\repair} and the proposed training strategies, we designed four variants: 
(1) \textit{Rand-R}, which replaces {\repair} with a randomly selected repair decision from the action space for states identified by {\monitor}; 
(2) \textit{Only SL}, representing the variant that directly employs {\tool} trained solely via the supervised learning stage, without reinforcement fine-tuning; 
(3) \textit{Only RL}, a variant that trains {\tool} purely through reinforcement learning;
(4) \textit{wo SR}, a variant that omits the step reward defined in Algorithm~\ref{algo:reft} (Lines 16-20). 
Table~\ref{tab:rq2-repair} shows comparative results.

(a) \textit{Adaptive Ability.}
We observe that directly using \textit{Only SL} achieves relatively good repair performance, with $\Delta E$ values of 75.5 for \textit{Roach} and 72.0 for \textit{Pylot}. 
Comparing \textit{Only SL} with \textit{Rand-R}, we observe a significant decline in repair performance with \textit{Rand-R} (75.5 vs. 63.5 for \textit{Roach} and 72.0 vs. 56.5 for \textit{Pylot}), along with unstable variations in \textit{\%Intensity}—specifically, a decrease for \textit{Roach} (0.98 vs. 0.89) but an increase for \textit{Pylot} (0.87 vs. 0.93). This demonstrates that arbitrarily modifying intervention intensity can negatively impact safety.
In contrast, when comparing {\tool} with \textit{Only SL}, {\tool} not only improves repair performance (85.0 vs. 75.5 for \textit{Roach} and 76.0 vs. 72.0 for \textit{Pylot}), but also reduces intervention intensity (\textit{\%Intensity}) (0.77 vs. 0.98 for \textit{Roach} and 0.72 vs. 0.87 for \textit{Pylot}).
These results demonstrate the adaptive capability of {\tool} in making repair decisions tailored to the driving context, effectively enhancing safety while minimizing unnecessary intervention.

(b) \textit{Training Strategies.}  
\textit{Only SL}, trained solely with supervised learning, maintains a baseline level of repair performance. However, when comparing \textit{Only RL} with \textit{Only SL}, we find that training \tool{} exclusively via Reinforcement Learning using the reward function defined in Algorithm~\ref{algo:reft} poses challenges in achieving satisfactory repair performance.
Specifically, \textit{Only RL} demonstrates the weakest overall repair performance, with notably low \textit{\%Fixed} (36.5 for \textit{Roach} and 32.0 for \textit{Pylot}) and $\Delta E$ values (30.5 for \textit{Roach} and 28.5 for \textit{Pylot}). 
This poor performance persists despite \textit{Only RL} achieving the lowest intervention intensity (\%Intensity of 0.19 for \textit{Roach} and 0.23 for \textit{Pylot}), indicating that pure RL tends to prioritize minimizing intervention, which does not necessarily lead to effective repairs. 
In contrast, fine-tuning \textit{Only SL} to obtain \tool{} significantly reduces intervention intensity (0.98 vs. 0.77 for \textit{Roach} and 0.87 vs. 0.72 for \textit{Pylot}), while simultaneously improving overall repair performance ($\Delta E$) (75.5 vs. 85.0 for \textit{Roach} and 72.0 vs. 76.0 for \textit{Pylot}). This demonstrates the effectiveness of ReFT in optimizing the repair process by refining repair decisions through exploration near the already effective solutions identified by \textit{Only SL}.
Additionally, \textit{wo SR}, which fine-tunes \textit{Only SL} without incorporating the step reward (Line 17 in Algorithm~\ref{algo:reft}), leads to an increase in \textit{\%Intensity} compared to \tool{} (1.02 vs. 0.72 for \textit{Pylot} and 0.83 vs. 0.77 for \textit{Roach}). This underscores the importance of step rewards in guiding \tool{} toward more optimal repair decisions, ensuring safety while minimizing disruptions to the original ADS motion.

\begin{center}
\vspace{-10pt}
\fcolorbox{black}{gray!10}{\parbox{\linewidth}{\textbf{Answer to RQ2}: \monitor{} effectively detects safety-critical states in advance, while \repair{} and the training strategies ensure adaptive and optimal repairs, enhancing safety with minimal impact on ADS driving.  
}}
\end{center}








\subsection{RQ3: Cost of \tool}
\begin{table}[!t]
    \centering
    \caption{Computation cost for processing one driving scene, measured in milliseconds (ms).}
    \vspace{-10pt}
    \small
    \resizebox{1.0\linewidth}{!}{
        \begin{tabular}{l|ccc|cc}
        \toprule
         \multirow{2.5}*{\textit{Method}} & \multicolumn{3}{c|}{Repair Mechanism} & \multicolumn{2}{c}{\textit{ADS}} \\ 
         \cmidrule(lr){2-4}\cmidrule(lr){5-6}
         & \tool & \textit{RuleRepair} & \textit{ModelRepair} & \textit{Roach} & \textit{Pylot} \\
         \midrule
         \textit{time (ms)} & 3.421 & 0.718 & 1.119 & 25.183 & 88.727 \\
        \bottomrule
        \end{tabular}
    }
    \label{tab:RQ3-performance}
    \vspace{-10pt}
\end{table}

Table~\ref{tab:RQ3-performance} presents the computational costs of \tool{}. Compared to existing repair baselines (\textit{RuleRepair} and \textit{ModelRepair}), \tool{} requires more time to process a single driving frame (3.421 ms vs. 0.487 ms for \textit{RuleRepair} and 1.119 ms for \textit{ModelRepair}).  
The primary reason for \tool{}'s higher computational cost is the use of a Transformer-based encoder, which accounts for most of the processing time. However, despite this overhead, \tool{} remains significantly faster than the original ADSs, which require 25.183 ms for \textit{Roach} and 88.727 ms for \textit{Pylot}. This ensures that \tool{} does not introduce excessive latency or hinder the processing of subsequent frames.  
Furthermore, \tool{} has a compact model size of only 17.26 MB, making it effective and efficient for deployment.



\begin{center}
\vspace{-10pt}
\fcolorbox{black}{gray!10}{\parbox{\linewidth}{\textbf{Answer to RQ3}: \tool demonstrates a significantly low computational cost, requiring only 3.421 ms to process one frame.
}}
\end{center}
\subsection{Threats to Validity}
Although \tool{} demonstrates a strong ability to ensure safety while minimizing its impact on the original ADS's driving behavior, certain threats remain to be validated.  
The first threat is that \tool{} assumes perfect perception results. Imperfect perception introduces uncertainty, making it challenging to distinguish whether violations originate from decision-making errors or perception inaccuracies. Since \tool{} is designed to address safety-critical violations caused by decision-making errors, we evaluate its effectiveness under perfect perception conditions to ensure a fair assessment by following previous studies~\cite{av_fuzzer,cheng2023behavexplor, sun2024redriver, sun2022lawbreaker}. The impact of imperfect perception will be investigated in future work. 
The second threat lies in the choice of action space and violation types. \tool{} focuses solely on collision avoidance by adjusting longitudinal movement through throttle-braking action space, similar to previous studies~\cite{coelingh2010collision, carla_behavior, grieser2020assuring}, without incorporating higher-level repair strategies such as overtaking maneuvers that require steering adjustments.  
While integrating higher-level repair strategies could enable \tool{} to address a broader range of decision-making faults, designing and training such strategies introduces significant complexity, which falls outside the scope of this work. Our results demonstrate that adjusting longitudinal movement alone is sufficient to prevent most collisions. We leave the exploration of an extended action space and the handling of more diverse violation types (e.g., non-optimal decision paths) as future work.
Finally, the choice of representation introduces certain threats. In this work, we utilize perception results in the form of attributes (i.e., location, heading, speed), which requires no additional conversion efforts and remains compatible with both module-based and end-to-end ADSs. However, alternative representations, such as Bird's-Eye-View, may impact the results. Exploring the influence of different perception representations is left for future work.

\section{Related Work}







\noindent\textbf{Autonomous Driving System Testing.} ADS testing is an important technique to evaluate ADSs by generating safety-critical scenarios. Existing online ADS testing studies are mainly divded into data-driven approaches \cite{zhang2023building,deng2022scenario,gambi2019generating,najm2013depiction,nitsche2017pre,roesener2016scenario, paardekooper2019automatic} and searching-based approaches \cite{cheng2023behavexplor,hildebrandt2023physcov,gambi2019automatically,han2021preliminary,av_fuzzer,icse_samota,tse_adfuzz,tang2021systematic,zhou2023specification,tang2021route,tang2021collision,huai2023doppelganger,wang2025moditector,cheng2024evaluating,tang2025moral}. 
Data-driven methods generate critical scenarios from real-world data, such as traffic reports~\cite{gambi2019generating,najm2013depiction,nitsche2017pre,zhang2023building}. 
Search-based methods use various technologies to search for safety-critical scenarios from the scenario space, such as guided fuzzing~\cite{cheng2023behavexplor, av_fuzzer, MDPFuzz_2022_issta, tian2022generating}, evolutionary algorithms~\cite{gambi2019automatically,han2021preliminary, tang2021collision,tang2021route,tang2021systematic,zhou2023specification,tian2022mosat}, metamorphic testing~\cite{han2020metamorphic}, surrogate models~\cite{icse_samota,tse_adfuzz}, reinforcement learning \cite{haq2023many, feng2023dense, lu2022learning} and reachability analysis~\cite{hildebrandt2023physcov,althoff2018automatic}. All these methods aim to generate critical scenarios for evaluating ADSs' safety-critical requirements, such as collision avoidance. In our paper, we employ search-based testing techniques to collect the repair verification benchmark and verify the robustness of \tool.


\noindent\textbf{Runtime Safety Guarantees in Autonomous Driving.} 
Ensuring runtime safety is crucial for developing safe and reliable ADSs. 
Existing studies can be divided into runtime monitor~\cite{watanabe2018runtime, stocco2020misbehaviour, stocco2022thirdeye} and runtime enforcement~\cite{grieser2020assuring, sun2024redriver, carla_behavior}. 
The runtime monitor aims to timely identify safety-critical driving scenes during online operations. Recent monitors~\cite{stocco2020misbehaviour,stocco2022thirdeye} leverage embeddings learned from neural networks to detect out-of-distribution data, effectively identifying unseen scenarios. However, these monitors struggle to properly handle violations in seen scenarios. 
Runtime enforcement~\cite{grieser2020assuring, sun2024redriver, carla_behavior} aims to detect driving states that violate predefined specifications, such as traffic rules~\cite{sun2024redriver} and the minimum distance required to avoid collisions with other participants~\cite{grieser2020assuring, carla_behavior}. Upon detecting such violations, the enforcement mechanism ensures that the ego vehicle adheres to the expected values defined in these specifications, which may lead to lower driving quality (i.e., comfort). To the best of our knowledge, there is currently no runtime safety guarantee system that ensures the online safety of ADSs while maintaining driving quality. In our paper, we explore a model-based repair framework designed to automatically and adaptively select a balanced repair decision, without requiring specific annotations.

\section{Conclusion}
In this paper, we propose an adaptive online decision repair method for autonomous driving systems (ADSs) that adaptively selects repair actions to mitigate violations while maintaining a low repair intensity.
Specifically, we design a learning-based framework, \tool, consisting of three main components: \encoder{}, \monitor{}, and \repair{}. 
\encoder{} utilizes a transformer-based encoder to extract high-level environmental representations, \monitor{} predicts the safety-critical score for each driving state, and \repair{} predicts a repaired action based on the environmental representation. \tool{} finally merges the predicted repaired action with the ADS action according to the safety-critical score.
A two-stage training pipeline is designed to achieve the goal of predicting effective repair actions with minimal intervention intensity and overcoming the limitation of lacking fine-grained annotations.
The experimental results demonstrate the effectiveness of \tool{} in repairing violations with less intensive repair actions while maintaining a robust ability to ensure runtime safety. 
Moreover, \tool{} introduces only minimal computational cost, facilitating easy deployment on existing ADS platforms without significant computational overhead.

\bibliographystyle{ACM-Reference-Format}
\bibliography{reference}


\begin{thebibliography}{66}


\ifx \showCODEN    \undefined \def \showCODEN     #1{\unskip}     \fi
\ifx \showISBNx    \undefined \def \showISBNx     #1{\unskip}     \fi
\ifx \showISBNxiii \undefined \def \showISBNxiii  #1{\unskip}     \fi
\ifx \showISSN     \undefined \def \showISSN      #1{\unskip}     \fi
\ifx \showLCCN     \undefined \def \showLCCN      #1{\unskip}     \fi
\ifx \shownote     \undefined \def \shownote      #1{#1}          \fi
\ifx \showarticletitle \undefined \def \showarticletitle #1{#1}   \fi
\ifx \showURL      \undefined \def \showURL       {\relax}        \fi
\providecommand\bibfield[2]{#2}
\providecommand\bibinfo[2]{#2}
\providecommand\natexlab[1]{#1}
\providecommand\showeprint[2][]{arXiv:#2}

\bibitem[Althoff and Lutz(2018)]%
        {althoff2018automatic}
\bibfield{author}{\bibinfo{person}{Matthias Althoff} {and} \bibinfo{person}{Sebastian Lutz}.} \bibinfo{year}{2018}\natexlab{}.
\newblock \showarticletitle{Automatic generation of safety-critical test scenarios for collision avoidance of road vehicles}. In \bibinfo{booktitle}{\emph{2018 IEEE Intelligent Vehicles Symposium (IV)}}. \bibinfo{publisher}{IEEE}, \bibinfo{address}{Changshu, Suzhou, China}, \bibinfo{pages}{1326--1333}.
\newblock


\bibitem[Baidu(2019)]%
        {apollo}
\bibfield{author}{\bibinfo{person}{Baidu}.} \bibinfo{year}{2019}\natexlab{}.
\newblock \bibinfo{title}{Apollo: Open Source Autonomous Driving}.
\newblock
\urldef\tempurl%
\url{https://github.com/ApolloAuto/apollo}
\showURL{%
\tempurl}


\bibitem[CARLA(2020)]%
        {carla_behavior}
\bibfield{author}{\bibinfo{person}{CARLA}.} \bibinfo{year}{2020}\natexlab{}.
\newblock \bibinfo{title}{Behavior Agent in Carla}.
\newblock


\bibitem[Chen et~al\mbox{.}(2019)]%
        {chen2019deep}
\bibfield{author}{\bibinfo{person}{Jianyu Chen}, \bibinfo{person}{Bodi Yuan}, {and} \bibinfo{person}{Masayoshi Tomizuka}.} \bibinfo{year}{2019}\natexlab{}.
\newblock \showarticletitle{Deep imitation learning for autonomous driving in generic urban scenarios with enhanced safety}. In \bibinfo{booktitle}{\emph{2019 IEEE/RSJ international conference on intelligent robots and systems (IROS)}}. IEEE, \bibinfo{pages}{2884--2890}.
\newblock


\bibitem[Cheng et~al\mbox{.}(2025)]%
        {cheng2024evaluating}
\bibfield{author}{\bibinfo{person}{Mingfei Cheng}, \bibinfo{person}{Xiaofei Xie}, \bibinfo{person}{Yuan Zhou}, \bibinfo{person}{Junjie Wang}, \bibinfo{person}{Guozhu Meng}, {and} \bibinfo{person}{Kairui Yang}.} \bibinfo{year}{2025}\natexlab{}.
\newblock \showarticletitle{{Decictor}: Towards Evaluating the Robustness of Decision-Making in Autonomous Driving Systems}. In \bibinfo{booktitle}{\emph{Proceedings of the 47th IEEE/ACM International Conference on Software Engineering (ICSE)}}. \bibinfo{pages}{651--651}.
\newblock


\bibitem[Cheng et~al\mbox{.}(2023)]%
        {cheng2023behavexplor}
\bibfield{author}{\bibinfo{person}{Mingfei Cheng}, \bibinfo{person}{Yuan Zhou}, {and} \bibinfo{person}{Xiaofei Xie}.} \bibinfo{year}{2023}\natexlab{}.
\newblock \showarticletitle{BehAVExplor: Behavior Diversity Guided Testing for Autonomous Driving Systems}. In \bibinfo{booktitle}{\emph{Proceedings of the 32nd ACM SIGSOFT International Symposium on Software Testing and Analysis}}. \bibinfo{pages}{488--500}.
\newblock


\bibitem[Coelingh et~al\mbox{.}(2010)]%
        {coelingh2010collision}
\bibfield{author}{\bibinfo{person}{Erik Coelingh}, \bibinfo{person}{Andreas Eidehall}, {and} \bibinfo{person}{Mattias Bengtsson}.} \bibinfo{year}{2010}\natexlab{}.
\newblock \showarticletitle{Collision warning with full auto brake and pedestrian detection-a practical example of automatic emergency braking}. In \bibinfo{booktitle}{\emph{13th International IEEE Conference on Intelligent Transportation Systems}}. IEEE, \bibinfo{pages}{155--160}.
\newblock


\bibitem[comma.ai(2022)]%
        {openpilot}
\bibfield{author}{\bibinfo{person}{comma.ai}.} \bibinfo{year}{2022}\natexlab{}.
\newblock \bibinfo{title}{{OpenPilot}: An open source driver assistance system}.
\newblock
\urldef\tempurl%
\url{https://github.com/commaai/openpilot}
\showURL{%
Retrieved Nov 7, 2022 from \tempurl}


\bibitem[Deng et~al\mbox{.}(2022)]%
        {deng2022scenario}
\bibfield{author}{\bibinfo{person}{Yao Deng}, \bibinfo{person}{Xi Zheng}, \bibinfo{person}{Mengshi Zhang}, \bibinfo{person}{Guannan Lou}, {and} \bibinfo{person}{Tianyi Zhang}.} \bibinfo{year}{2022}\natexlab{}.
\newblock \showarticletitle{Scenario-based test reduction and prioritization for multi-module autonomous driving systems}. In \bibinfo{booktitle}{\emph{Proceedings of the 30th ACM Joint European Software Engineering Conference and Symposium on the Foundations of Software Engineering}}. \bibinfo{pages}{82--93}.
\newblock


\bibitem[Devlin et~al\mbox{.}(2019)]%
        {devlin2019bert}
\bibfield{author}{\bibinfo{person}{Jacob Devlin}, \bibinfo{person}{Ming-Wei Chang}, \bibinfo{person}{Kenton Lee}, {and} \bibinfo{person}{Kristina Toutanova}.} \bibinfo{year}{2019}\natexlab{}.
\newblock \showarticletitle{Bert: Pre-training of deep bidirectional transformers for language understanding}. In \bibinfo{booktitle}{\emph{Proceedings of the 2019 conference of the North American chapter of the association for computational linguistics: human language technologies, volume 1 (long and short papers)}}. \bibinfo{pages}{4171--4186}.
\newblock


\bibitem[Dosovitskiy et~al\mbox{.}(2020)]%
        {dosovitskiy2020image}
\bibfield{author}{\bibinfo{person}{Alexey Dosovitskiy}, \bibinfo{person}{Lucas Beyer}, \bibinfo{person}{Alexander Kolesnikov}, \bibinfo{person}{Dirk Weissenborn}, \bibinfo{person}{Xiaohua Zhai}, \bibinfo{person}{Thomas Unterthiner}, \bibinfo{person}{Mostafa Dehghani}, \bibinfo{person}{Matthias Minderer}, \bibinfo{person}{Georg Heigold}, \bibinfo{person}{Sylvain Gelly}, {et~al\mbox{.}}} \bibinfo{year}{2020}\natexlab{}.
\newblock \showarticletitle{An image is worth 16x16 words: Transformers for image recognition at scale}.
\newblock \bibinfo{journal}{\emph{arXiv preprint arXiv:2010.11929}} (\bibinfo{year}{2020}).
\newblock


\bibitem[Dosovitskiy et~al\mbox{.}(2017)]%
        {dosovitskiy2017carla}
\bibfield{author}{\bibinfo{person}{Alexey Dosovitskiy}, \bibinfo{person}{German Ros}, \bibinfo{person}{Felipe Codevilla}, \bibinfo{person}{Antonio Lopez}, {and} \bibinfo{person}{Vladlen Koltun}.} \bibinfo{year}{2017}\natexlab{}.
\newblock \showarticletitle{{CARLA}: {An} Open Urban Driving Simulator}. In \bibinfo{booktitle}{\emph{Proceedings of the 1st Annual Conference on Robot Learning}} \emph{(\bibinfo{series}{Proceedings of Machine Learning Research}, Vol.~\bibinfo{volume}{78})}, \bibfield{editor}{\bibinfo{person}{Sergey Levine}, \bibinfo{person}{Vincent Vanhoucke}, {and} \bibinfo{person}{Ken Goldberg}} (Eds.). \bibinfo{publisher}{PMLR}, \bibinfo{address}{California, USA}, \bibinfo{pages}{1--16}.
\newblock


\bibitem[Feng et~al\mbox{.}(2023)]%
        {feng2023dense}
\bibfield{author}{\bibinfo{person}{Shuo Feng}, \bibinfo{person}{Haowei Sun}, \bibinfo{person}{Xintao Yan}, \bibinfo{person}{Haojie Zhu}, \bibinfo{person}{Zhengxia Zou}, \bibinfo{person}{Shengyin Shen}, {and} \bibinfo{person}{Henry~X Liu}.} \bibinfo{year}{2023}\natexlab{}.
\newblock \showarticletitle{Dense reinforcement learning for safety validation of autonomous vehicles}.
\newblock \bibinfo{journal}{\emph{Nature}} \bibinfo{volume}{615}, \bibinfo{number}{7953} (\bibinfo{year}{2023}), \bibinfo{pages}{620--627}.
\newblock


\bibitem[Gambi et~al\mbox{.}(2019a)]%
        {gambi2019generating}
\bibfield{author}{\bibinfo{person}{Alessio Gambi}, \bibinfo{person}{Tri Huynh}, {and} \bibinfo{person}{Gordon Fraser}.} \bibinfo{year}{2019}\natexlab{a}.
\newblock \showarticletitle{Generating effective test cases for self-driving cars from police reports}. In \bibinfo{booktitle}{\emph{Proceedings of the 2019 27th ACM Joint Meeting on European Software Engineering Conference and Symposium on the Foundations of Software Engineering}}. \bibinfo{publisher}{ACM}, \bibinfo{address}{Tallinn Estonia}, \bibinfo{pages}{257--267}.
\newblock


\bibitem[Gambi et~al\mbox{.}(2019b)]%
        {gambi2019automatically}
\bibfield{author}{\bibinfo{person}{Alessio Gambi}, \bibinfo{person}{Marc Mueller}, {and} \bibinfo{person}{Gordon Fraser}.} \bibinfo{year}{2019}\natexlab{b}.
\newblock \showarticletitle{Automatically testing self-driving cars with search-based procedural content generation}. In \bibinfo{booktitle}{\emph{Proceedings of the 28th ACM SIGSOFT International Symposium on Software Testing and Analysis}}. \bibinfo{publisher}{ACM}, \bibinfo{address}{Beijing, China}, \bibinfo{pages}{318--328}.
\newblock


\bibitem[Garcia et~al\mbox{.}(2020)]%
        {garcia2020comprehensive}
\bibfield{author}{\bibinfo{person}{Joshua Garcia}, \bibinfo{person}{Yang Feng}, \bibinfo{person}{Junjie Shen}, \bibinfo{person}{Sumaya Almanee}, \bibinfo{person}{Yuan Xia}, {and} \bibinfo{person}{Qi~Alfred Chen}.} \bibinfo{year}{2020}\natexlab{}.
\newblock \showarticletitle{A comprehensive study of autonomous vehicle bugs}. In \bibinfo{booktitle}{\emph{Proceedings of the ACM/IEEE 42nd International Conference on Software Engineering}}. \bibinfo{publisher}{IEEE}, \bibinfo{address}{Seoul, South Korea}, \bibinfo{pages}{385--396}.
\newblock


\bibitem[Gog et~al\mbox{.}(2021)]%
        {gog2021pylot}
\bibfield{author}{\bibinfo{person}{Ionel Gog}, \bibinfo{person}{Sukrit Kalra}, \bibinfo{person}{Peter Schafhalter}, \bibinfo{person}{Matthew~A Wright}, \bibinfo{person}{Joseph~E Gonzalez}, {and} \bibinfo{person}{Ion Stoica}.} \bibinfo{year}{2021}\natexlab{}.
\newblock \showarticletitle{Pylot: A modular platform for exploring latency-accuracy tradeoffs in autonomous vehicles}. In \bibinfo{booktitle}{\emph{2021 IEEE International Conference on Robotics and Automation (ICRA)}}. IEEE, \bibinfo{pages}{8806--8813}.
\newblock


\bibitem[Grieser et~al\mbox{.}(2020)]%
        {grieser2020assuring}
\bibfield{author}{\bibinfo{person}{J{\"o}rg Grieser}, \bibinfo{person}{Meng Zhang}, \bibinfo{person}{Tim Warnecke}, {and} \bibinfo{person}{Andreas Rausch}.} \bibinfo{year}{2020}\natexlab{}.
\newblock \showarticletitle{Assuring the safety of end-to-end learning-based autonomous driving through runtime monitoring}. In \bibinfo{booktitle}{\emph{2020 23rd Euromicro Conference on Digital System Design (DSD)}}. IEEE, \bibinfo{pages}{476--483}.
\newblock


\bibitem[Han and Zhou(2020)]%
        {han2020metamorphic}
\bibfield{author}{\bibinfo{person}{Jia~Cheng Han} {and} \bibinfo{person}{Zhi~Quan Zhou}.} \bibinfo{year}{2020}\natexlab{}.
\newblock \showarticletitle{Metamorphic fuzz testing of autonomous vehicles}. In \bibinfo{booktitle}{\emph{Proceedings of the IEEE/ACM 42nd International Conference on Software Engineering Workshops}}. \bibinfo{pages}{380--385}.
\newblock


\bibitem[Han et~al\mbox{.}(2021)]%
        {han2021preliminary}
\bibfield{author}{\bibinfo{person}{Seunghee Han}, \bibinfo{person}{Jaeuk Kim}, \bibinfo{person}{Geon Kim}, \bibinfo{person}{Jaemin Cho}, \bibinfo{person}{Jiin Kim}, {and} \bibinfo{person}{Shin Yoo}.} \bibinfo{year}{2021}\natexlab{}.
\newblock \showarticletitle{Preliminary evaluation of path-aware crossover operators for search-based test data generation for autonomous driving}. In \bibinfo{booktitle}{\emph{2021 IEEE/ACM 14th International Workshop on Search-Based Software Testing (SBST)}}. \bibinfo{publisher}{IEEE}, \bibinfo{address}{Madrid, Spain}, \bibinfo{pages}{44--47}.
\newblock


\bibitem[Haq et~al\mbox{.}(2022)]%
        {icse_samota}
\bibfield{author}{\bibinfo{person}{Fitash~Ul Haq}, \bibinfo{person}{Donghwan Shin}, {and} \bibinfo{person}{Lionel Briand}.} \bibinfo{year}{2022}\natexlab{}.
\newblock \showarticletitle{Efficient online testing for DNN-enabled systems using surrogate-assisted and many-objective optimization}. In \bibinfo{booktitle}{\emph{Proceedings of the 44th International Conference on Software Engineering}}. \bibinfo{publisher}{IEEE}, \bibinfo{address}{Pittsburgh Pennsylvania}, \bibinfo{pages}{811--822}.
\newblock


\bibitem[Haq et~al\mbox{.}(2023)]%
        {haq2023many}
\bibfield{author}{\bibinfo{person}{Fitash~Ul Haq}, \bibinfo{person}{Donghwan Shin}, {and} \bibinfo{person}{Lionel~C Briand}.} \bibinfo{year}{2023}\natexlab{}.
\newblock \showarticletitle{Many-objective reinforcement learning for online testing of dnn-enabled systems}. In \bibinfo{booktitle}{\emph{2023 IEEE/ACM 45th International Conference on Software Engineering (ICSE)}}. IEEE, \bibinfo{pages}{1814--1826}.
\newblock


\bibitem[Hildebrandt et~al\mbox{.}(2023)]%
        {hildebrandt2023physcov}
\bibfield{author}{\bibinfo{person}{Carl Hildebrandt}, \bibinfo{person}{Meriel von Stein}, {and} \bibinfo{person}{Sebastian Elbaum}.} \bibinfo{year}{2023}\natexlab{}.
\newblock \showarticletitle{PhysCov: Physical Test Coverage for Autonomous Vehicles}. In \bibinfo{booktitle}{\emph{Proceedings of the 32nd ACM SIGSOFT International Symposium on Software Testing and Analysis}}. \bibinfo{pages}{449--461}.
\newblock


\bibitem[Hong et~al\mbox{.}(2020)]%
        {hong2020avguardian}
\bibfield{author}{\bibinfo{person}{David~Ke Hong}, \bibinfo{person}{John Kloosterman}, \bibinfo{person}{Yuqi Jin}, \bibinfo{person}{Yulong Cao}, \bibinfo{person}{Qi~Alfred Chen}, \bibinfo{person}{Scott Mahlke}, {and} \bibinfo{person}{Z~Morley Mao}.} \bibinfo{year}{2020}\natexlab{}.
\newblock \showarticletitle{AVGuardian: Detecting and mitigating publish-subscribe overprivilege for autonomous vehicle systems}. In \bibinfo{booktitle}{\emph{2020 IEEE European Symposium on Security and Privacy (EuroS\&P)}}. IEEE, \bibinfo{pages}{445--459}.
\newblock


\bibitem[Hu et~al\mbox{.}(2023)]%
        {hu2023_uniad}
\bibfield{author}{\bibinfo{person}{Yihan Hu}, \bibinfo{person}{Jiazhi Yang}, \bibinfo{person}{Li Chen}, \bibinfo{person}{Keyu Li}, \bibinfo{person}{Chonghao Sima}, \bibinfo{person}{Xizhou Zhu}, \bibinfo{person}{Siqi Chai}, \bibinfo{person}{Senyao Du}, \bibinfo{person}{Tianwei Lin}, \bibinfo{person}{Wenhai Wang}, \bibinfo{person}{Lewei Lu}, \bibinfo{person}{Xiaosong Jia}, \bibinfo{person}{Qiang Liu}, \bibinfo{person}{Jifeng Dai}, \bibinfo{person}{Yu Qiao}, {and} \bibinfo{person}{Hongyang Li}.} \bibinfo{year}{2023}\natexlab{}.
\newblock \showarticletitle{Planning-oriented Autonomous Driving}. In \bibinfo{booktitle}{\emph{Proceedings of the IEEE/CVF Conference on Computer Vision and Pattern Recognition}}.
\newblock


\bibitem[Huai et~al\mbox{.}(2023a)]%
        {huai2023sceno}
\bibfield{author}{\bibinfo{person}{Yuqi Huai}, \bibinfo{person}{Sumaya Almanee}, \bibinfo{person}{Yuntianyi Chen}, \bibinfo{person}{Xiafa Wu}, \bibinfo{person}{Qi~Alfred Chen}, {and} \bibinfo{person}{Joshua Garcia}.} \bibinfo{year}{2023}\natexlab{a}.
\newblock \showarticletitle{sceno RITA: Generating Diverse, Fully-Mutable, Test Scenarios for Autonomous Vehicle Planning}.
\newblock \bibinfo{journal}{\emph{IEEE Transactions on Software Engineering}} (\bibinfo{year}{2023}).
\newblock


\bibitem[Huai et~al\mbox{.}(2023b)]%
        {huai2023doppelganger}
\bibfield{author}{\bibinfo{person}{Yuqi Huai}, \bibinfo{person}{Yuntianyi Chen}, \bibinfo{person}{Sumaya Almanee}, \bibinfo{person}{Tuan Ngo}, \bibinfo{person}{Xiang Liao}, \bibinfo{person}{Ziwen Wan}, \bibinfo{person}{Qi~Alfred Chen}, {and} \bibinfo{person}{Joshua Garcia}.} \bibinfo{year}{2023}\natexlab{b}.
\newblock \showarticletitle{Doppelg{\"a}nger test generation for revealing bugs in autonomous driving software}. In \bibinfo{booktitle}{\emph{2023 IEEE/ACM 45th International Conference on Software Engineering (ICSE)}}. IEEE, \bibinfo{pages}{2591--2603}.
\newblock


\bibitem[Kato et~al\mbox{.}(2018)]%
        {autoware}
\bibfield{author}{\bibinfo{person}{Shinpei Kato}, \bibinfo{person}{Shota Tokunaga}, \bibinfo{person}{Yuya Maruyama}, \bibinfo{person}{Seiya Maeda}, \bibinfo{person}{Manato Hirabayashi}, \bibinfo{person}{Yuki Kitsukawa}, \bibinfo{person}{Abraham Monrroy}, \bibinfo{person}{Tomohito Ando}, \bibinfo{person}{Yusuke Fujii}, {and} \bibinfo{person}{Takuya Azumi}.} \bibinfo{year}{2018}\natexlab{}.
\newblock \showarticletitle{Autoware on board: Enabling autonomous vehicles with embedded systems}. In \bibinfo{booktitle}{\emph{2018 ACM/IEEE 9th International Conference on Cyber-Physical Systems (ICCPS)}}. IEEE, \bibinfo{pages}{287--296}.
\newblock


\bibitem[Kim et~al\mbox{.}(2022)]%
        {css_drivefuzzer}
\bibfield{author}{\bibinfo{person}{Seulbae Kim}, \bibinfo{person}{Major Liu}, \bibinfo{person}{Junghwan"~John" Rhee}, \bibinfo{person}{Yuseok Jeon}, \bibinfo{person}{Yonghwi Kwon}, {and} \bibinfo{person}{Chung~Hwan Kim}.} \bibinfo{year}{2022}\natexlab{}.
\newblock \showarticletitle{DriveFuzz: Discovering Autonomous Driving Bugs through Driving Quality-Guided Fuzzing}. In \bibinfo{booktitle}{\emph{Proceedings of the 2022 ACM SIGSAC Conference on Computer and Communications Security}}. \bibinfo{publisher}{ACM}, \bibinfo{address}{Los Angeles, CA, USA}, \bibinfo{pages}{1753--1767}.
\newblock


\bibitem[Lee and Lee(2013)]%
        {lee2013study}
\bibfield{author}{\bibinfo{person}{Jeong~Keun Lee} {and} \bibinfo{person}{Kang~Wook Lee}.} \bibinfo{year}{2013}\natexlab{}.
\newblock \showarticletitle{Study on Effectiveness of Pre-Crash Active Seatbelt Using Real Time Controlled Simulation}. In \bibinfo{booktitle}{\emph{23rd International Technical Conference on the Enhanced Safety of Vehicles (ESV) National Highway Traffic Safety Administration}}.
\newblock


\bibitem[Li et~al\mbox{.}(2020)]%
        {av_fuzzer}
\bibfield{author}{\bibinfo{person}{Guanpeng Li}, \bibinfo{person}{Yiran Li}, \bibinfo{person}{Saurabh Jha}, \bibinfo{person}{Timothy Tsai}, \bibinfo{person}{Michael Sullivan}, \bibinfo{person}{Siva Kumar~Sastry Hari}, \bibinfo{person}{Zbigniew Kalbarczyk}, {and} \bibinfo{person}{Ravishankar Iyer}.} \bibinfo{year}{2020}\natexlab{}.
\newblock \showarticletitle{{AV-FUZZER}: Finding safety violations in autonomous driving systems}. In \bibinfo{booktitle}{\emph{2020 IEEE 31st International Symposium on Software Reliability Engineering (ISSRE)}}. \bibinfo{publisher}{IEEE}, \bibinfo{address}{Coimbra, Portugal}, \bibinfo{pages}{25--36}.
\newblock


\bibitem[Liu et~al\mbox{.}(2023)]%
        {liu2023bevfusion}
\bibfield{author}{\bibinfo{person}{Zhijian Liu}, \bibinfo{person}{Haotian Tang}, \bibinfo{person}{Alexander Amini}, \bibinfo{person}{Xinyu Yang}, \bibinfo{person}{Huizi Mao}, \bibinfo{person}{Daniela~L Rus}, {and} \bibinfo{person}{Song Han}.} \bibinfo{year}{2023}\natexlab{}.
\newblock \showarticletitle{Bevfusion: Multi-task multi-sensor fusion with unified bird's-eye view representation}. In \bibinfo{booktitle}{\emph{2023 IEEE international conference on robotics and automation (ICRA)}}. IEEE, \bibinfo{pages}{2774--2781}.
\newblock


\bibitem[Lu et~al\mbox{.}(2022)]%
        {lu2022learning}
\bibfield{author}{\bibinfo{person}{Chengjie Lu}, \bibinfo{person}{Yize Shi}, \bibinfo{person}{Huihui Zhang}, \bibinfo{person}{Man Zhang}, \bibinfo{person}{Tiexin Wang}, \bibinfo{person}{Tao Yue}, {and} \bibinfo{person}{Shaukat Ali}.} \bibinfo{year}{2022}\natexlab{}.
\newblock \showarticletitle{Learning configurations of operating environment of autonomous vehicles to maximize their collisions}.
\newblock \bibinfo{journal}{\emph{IEEE Transactions on Software Engineering}} \bibinfo{volume}{49}, \bibinfo{number}{1} (\bibinfo{year}{2022}), \bibinfo{pages}{384--402}.
\newblock


\bibitem[Mauritz et~al\mbox{.}(2016)]%
        {mauritz2016assuring}
\bibfield{author}{\bibinfo{person}{Malte Mauritz}, \bibinfo{person}{Falk Howar}, {and} \bibinfo{person}{Andreas Rausch}.} \bibinfo{year}{2016}\natexlab{}.
\newblock \showarticletitle{Assuring the safety of advanced driver assistance systems through a combination of simulation and runtime monitoring}. In \bibinfo{booktitle}{\emph{Leveraging Applications of Formal Methods, Verification and Validation: Discussion, Dissemination, Applications: 7th International Symposium, ISoLA 2016, Imperial, Corfu, Greece, October 10-14, 2016, Proceedings, Part II 7}}. Springer, \bibinfo{pages}{672--687}.
\newblock


\bibitem[Monperrus(2018)]%
        {monperrus2018automatic}
\bibfield{author}{\bibinfo{person}{Martin Monperrus}.} \bibinfo{year}{2018}\natexlab{}.
\newblock \showarticletitle{Automatic software repair: A bibliography}.
\newblock \bibinfo{journal}{\emph{ACM Computing Surveys (CSUR)}} \bibinfo{volume}{51}, \bibinfo{number}{1} (\bibinfo{year}{2018}), \bibinfo{pages}{1--24}.
\newblock


\bibitem[Najm et~al\mbox{.}(2007)]%
        {najm2007pre}
\bibfield{author}{\bibinfo{person}{Wassim~G Najm}, \bibinfo{person}{John~D Smith}, \bibinfo{person}{Mikio Yanagisawa}, {et~al\mbox{.}}} \bibinfo{year}{2007}\natexlab{}.
\newblock \bibinfo{booktitle}{\emph{Pre-crash scenario typology for crash avoidance research}}.
\newblock \bibinfo{type}{{T}echnical {R}eport}. \bibinfo{institution}{United States. Department of Transportation. National Highway Traffic Safety}.
\newblock


\bibitem[Najm et~al\mbox{.}(2013)]%
        {najm2013depiction}
\bibfield{author}{\bibinfo{person}{Wassim~G Najm}, \bibinfo{person}{Samuel Toma}, \bibinfo{person}{John Brewer}, {et~al\mbox{.}}} \bibinfo{year}{2013}\natexlab{}.
\newblock \bibinfo{booktitle}{\emph{Depiction of priority light-vehicle pre-crash scenarios for safety applications based on vehicle-to-vehicle communications}}.
\newblock \bibinfo{type}{{T}echnical {R}eport} DOT HS 811 732. \bibinfo{institution}{National Highway Traffic Safety Administration, U.S. Department of Transportation}, \bibinfo{address}{Washington, DC}.
\newblock


\bibitem[Nitsche et~al\mbox{.}(2017)]%
        {nitsche2017pre}
\bibfield{author}{\bibinfo{person}{Philippe Nitsche}, \bibinfo{person}{Pete Thomas}, \bibinfo{person}{Rainer Stuetz}, {and} \bibinfo{person}{Ruth Welsh}.} \bibinfo{year}{2017}\natexlab{}.
\newblock \showarticletitle{Pre-crash scenarios at road junctions: A clustering method for car crash data}.
\newblock \bibinfo{journal}{\emph{Accident Analysis \& Prevention}}  \bibinfo{volume}{107} (\bibinfo{year}{2017}), \bibinfo{pages}{137--151}.
\newblock


\bibitem[Paardekooper et~al\mbox{.}(2019)]%
        {paardekooper2019automatic}
\bibfield{author}{\bibinfo{person}{Jan-Pieter Paardekooper}, \bibinfo{person}{S Montfort}, \bibinfo{person}{Jeroen Manders}, \bibinfo{person}{Jorrit Goos}, \bibinfo{person}{E~de Gelder}, \bibinfo{person}{O Camp}, \bibinfo{person}{O Bracquemond}, {and} \bibinfo{person}{Gildas Thiolon}.} \bibinfo{year}{2019}\natexlab{}.
\newblock \showarticletitle{Automatic identification of critical scenarios in a public dataset of 6000 km of public-road driving}. In \bibinfo{booktitle}{\emph{26th International Technical Conference on the Enhanced Safety of Vehicles (ESV)}}. \bibinfo{publisher}{Mira Smart}, \bibinfo{address}{Eindhoven, Netherlands}.
\newblock


\bibitem[Pang et~al\mbox{.}(2022)]%
        {MDPFuzz_2022_issta}
\bibfield{author}{\bibinfo{person}{Qi Pang}, \bibinfo{person}{Yuanyuan Yuan}, {and} \bibinfo{person}{Shuai Wang}.} \bibinfo{year}{2022}\natexlab{}.
\newblock \showarticletitle{MDPFuzz: Testing Models Solving Markov Decision Processes}. In \bibinfo{booktitle}{\emph{Proceedings of the 31st ACM SIGSOFT International Symposium on Software Testing and Analysis}} (Virtual, South Korea) \emph{(\bibinfo{series}{ISSTA 2022})}. \bibinfo{publisher}{Association for Computing Machinery}, \bibinfo{address}{New York, NY, USA}, \bibinfo{pages}{378–390}.
\newblock
\showISBNx{9781450393799}
\href{https://doi.org/10.1145/3533767.3534388}{doi:\nolinkurl{10.1145/3533767.3534388}}


\bibitem[Renz et~al\mbox{.}(2023)]%
        {renz2023plant}
\bibfield{author}{\bibinfo{person}{Katrin Renz}, \bibinfo{person}{Kashyap Chitta}, \bibinfo{person}{Otniel-Bogdan Mercea}, \bibinfo{person}{A~Sophia Koepke}, \bibinfo{person}{Zeynep Akata}, {and} \bibinfo{person}{Andreas Geiger}.} \bibinfo{year}{2023}\natexlab{}.
\newblock \showarticletitle{PlanT: Explainable Planning Transformers via Object-Level Representations}. In \bibinfo{booktitle}{\emph{Conference on Robot Learning}}. PMLR, \bibinfo{pages}{459--470}.
\newblock


\bibitem[Roesener et~al\mbox{.}(2016)]%
        {roesener2016scenario}
\bibfield{author}{\bibinfo{person}{Christian Roesener}, \bibinfo{person}{Felix Fahrenkrog}, \bibinfo{person}{Axel Uhlig}, {and} \bibinfo{person}{Lutz Eckstein}.} \bibinfo{year}{2016}\natexlab{}.
\newblock \showarticletitle{A scenario-based assessment approach for automated driving by using time series classification of human-driving behaviour}. In \bibinfo{booktitle}{\emph{2016 IEEE 19th international conference on intelligent transportation systems (ITSC)}}. \bibinfo{publisher}{IEEE}, \bibinfo{address}{Rio de Janeiro, Brazil}, \bibinfo{pages}{1360--1365}.
\newblock


\bibitem[Shankar et~al\mbox{.}(2020)]%
        {shankar2020formal}
\bibfield{author}{\bibinfo{person}{Saumya Shankar}, \bibinfo{person}{VR Ujwal}, \bibinfo{person}{Srinivas Pinisetty}, {and} \bibinfo{person}{Partha~S Roop}.} \bibinfo{year}{2020}\natexlab{}.
\newblock \showarticletitle{Formal Runtime Monitoring Approaches for Autonomous Vehicles.}
\newblock \bibinfo{journal}{\emph{OVERLAY}}  \bibinfo{volume}{20} (\bibinfo{year}{2020}), \bibinfo{pages}{89--94}.
\newblock


\bibitem[Stocco et~al\mbox{.}(2022)]%
        {stocco2022thirdeye}
\bibfield{author}{\bibinfo{person}{Andrea Stocco}, \bibinfo{person}{Paulo~J Nunes}, \bibinfo{person}{Marcelo d'Amorim}, {and} \bibinfo{person}{Paolo Tonella}.} \bibinfo{year}{2022}\natexlab{}.
\newblock \showarticletitle{Thirdeye: Attention maps for safe autonomous driving systems}. In \bibinfo{booktitle}{\emph{Proceedings of the 37th IEEE/ACM International Conference on Automated Software Engineering}}. \bibinfo{pages}{1--12}.
\newblock


\bibitem[Stocco et~al\mbox{.}(2020)]%
        {stocco2020misbehaviour}
\bibfield{author}{\bibinfo{person}{Andrea Stocco}, \bibinfo{person}{Michael Weiss}, \bibinfo{person}{Marco Calzana}, {and} \bibinfo{person}{Paolo Tonella}.} \bibinfo{year}{2020}\natexlab{}.
\newblock \showarticletitle{Misbehaviour prediction for autonomous driving systems}. In \bibinfo{booktitle}{\emph{Proceedings of the ACM/IEEE 42nd international conference on software engineering}}. \bibinfo{pages}{359--371}.
\newblock


\bibitem[Sun et~al\mbox{.}(2022)]%
        {sun2022lawbreaker}
\bibfield{author}{\bibinfo{person}{Yang Sun}, \bibinfo{person}{Christopher~M Poskitt}, \bibinfo{person}{Jun Sun}, \bibinfo{person}{Yuqi Chen}, {and} \bibinfo{person}{Zijiang Yang}.} \bibinfo{year}{2022}\natexlab{}.
\newblock \showarticletitle{LawBreaker: An approach for specifying traffic laws and fuzzing autonomous vehicles}. In \bibinfo{booktitle}{\emph{Proceedings of the 37th IEEE/ACM International Conference on Automated Software Engineering}}. \bibinfo{pages}{1--12}.
\newblock


\bibitem[Sun et~al\mbox{.}(2025)]%
        {sun2025fixdrive}
\bibfield{author}{\bibinfo{person}{Yang Sun}, \bibinfo{person}{Christopher~M Poskitt}, \bibinfo{person}{Kun Wang}, {and} \bibinfo{person}{Jun Sun}.} \bibinfo{year}{2025}\natexlab{}.
\newblock \showarticletitle{FixDrive: Automatically Repairing Autonomous Vehicle Driving Behaviour for \$0.08 per Violation}.
\newblock \bibinfo{journal}{\emph{arXiv preprint arXiv:2502.08260}} (\bibinfo{year}{2025}).
\newblock


\bibitem[Sun et~al\mbox{.}(2024)]%
        {sun2024redriver}
\bibfield{author}{\bibinfo{person}{Yang Sun}, \bibinfo{person}{Christopher~M Poskitt}, \bibinfo{person}{Xiaodong Zhang}, {and} \bibinfo{person}{Jun Sun}.} \bibinfo{year}{2024}\natexlab{}.
\newblock \showarticletitle{REDriver: Runtime Enforcement for Autonomous Vehicles}. In \bibinfo{booktitle}{\emph{Proceedings of the IEEE/ACM 46th International Conference on Software Engineering}}. \bibinfo{pages}{1--12}.
\newblock


\bibitem[Tang et~al\mbox{.}(2025)]%
        {tang2025moral}
\bibfield{author}{\bibinfo{person}{Wenbing Tang}, \bibinfo{person}{Mingfei Cheng}, \bibinfo{person}{Yuan Zhou}, {and} \bibinfo{person}{Yang Liu}.} \bibinfo{year}{2025}\natexlab{}.
\newblock \showarticletitle{Moral Testing of Autonomous Driving Systems}.
\newblock \bibinfo{journal}{\emph{arXiv preprint arXiv:2505.03683}} (\bibinfo{year}{2025}).
\newblock


\bibitem[Tang et~al\mbox{.}(2021a)]%
        {tang2021collision}
\bibfield{author}{\bibinfo{person}{Yun Tang}, \bibinfo{person}{Yuan Zhou}, \bibinfo{person}{Yang Liu}, \bibinfo{person}{Jun Sun}, {and} \bibinfo{person}{Gang Wang}.} \bibinfo{year}{2021}\natexlab{a}.
\newblock \showarticletitle{Collision avoidance testing for autonomous driving systems on complete maps}. In \bibinfo{booktitle}{\emph{2021 IEEE Intelligent Vehicles Symposium (IV)}}. \bibinfo{publisher}{IEEE}, \bibinfo{address}{Nagoya, Japan}, \bibinfo{pages}{179--185}.
\newblock


\bibitem[Tang et~al\mbox{.}(2021b)]%
        {tang2021route}
\bibfield{author}{\bibinfo{person}{Yun Tang}, \bibinfo{person}{Yuan Zhou}, \bibinfo{person}{Fenghua Wu}, \bibinfo{person}{Yang Liu}, \bibinfo{person}{Jun Sun}, \bibinfo{person}{Wuling Huang}, {and} \bibinfo{person}{Gang Wang}.} \bibinfo{year}{2021}\natexlab{b}.
\newblock \showarticletitle{Route coverage testing for autonomous vehicles via map modeling}. In \bibinfo{booktitle}{\emph{2021 IEEE International Conference on Robotics and Automation (ICRA)}}. \bibinfo{publisher}{IEEE}, \bibinfo{address}{Xi'an, China}, \bibinfo{pages}{11450--11456}.
\newblock


\bibitem[Tang et~al\mbox{.}(2021c)]%
        {tang2021systematic}
\bibfield{author}{\bibinfo{person}{Yun Tang}, \bibinfo{person}{Yuan Zhou}, \bibinfo{person}{Tianwei Zhang}, \bibinfo{person}{Fenghua Wu}, \bibinfo{person}{Yang Liu}, {and} \bibinfo{person}{Gang Wang}.} \bibinfo{year}{2021}\natexlab{c}.
\newblock \showarticletitle{Systematic testing of autonomous driving systems using map topology-based scenario classification}. In \bibinfo{booktitle}{\emph{Proceedings of the 36th IEEE/ACM International Conference on Automated Software Engineering (ASE)}}. \bibinfo{publisher}{IEEE}, \bibinfo{address}{Melbourne, Australia}, \bibinfo{pages}{1342--1346}.
\newblock


\bibitem[Tesla(2024)]%
        {tesla_fsd}
\bibfield{author}{\bibinfo{person}{Tesla}.} \bibinfo{year}{2024}\natexlab{}.
\newblock \bibinfo{title}{Autopilot and Full Self-Driving}.
\newblock


\bibitem[Thorn et~al\mbox{.}(2018)]%
        {thorn2018framework}
\bibfield{author}{\bibinfo{person}{Eric Thorn}, \bibinfo{person}{Shawn~C Kimmel}, \bibinfo{person}{Michelle Chaka}, \bibinfo{person}{Booz~Allen Hamilton}, {et~al\mbox{.}}} \bibinfo{year}{2018}\natexlab{}.
\newblock \bibinfo{booktitle}{\emph{A framework for automated driving system testable cases and scenarios}}.
\newblock \bibinfo{type}{{T}echnical {R}eport}. \bibinfo{institution}{United States. Department of Transportation. National Highway Traffic Safety}.
\newblock


\bibitem[Tian et~al\mbox{.}(2022a)]%
        {tian2022mosat}
\bibfield{author}{\bibinfo{person}{Haoxiang Tian}, \bibinfo{person}{Yan Jiang}, \bibinfo{person}{Guoquan Wu}, \bibinfo{person}{Jiren Yan}, \bibinfo{person}{Jun Wei}, \bibinfo{person}{Wei Chen}, \bibinfo{person}{Shuo Li}, {and} \bibinfo{person}{Dan Ye}.} \bibinfo{year}{2022}\natexlab{a}.
\newblock \showarticletitle{MOSAT: finding safety violations of autonomous driving systems using multi-objective genetic algorithm}. In \bibinfo{booktitle}{\emph{Proceedings of the 30th ACM Joint European Software Engineering Conference and Symposium on the Foundations of Software Engineering}}. \bibinfo{pages}{94--106}.
\newblock


\bibitem[Tian et~al\mbox{.}(2022b)]%
        {tian2022generating}
\bibfield{author}{\bibinfo{person}{Haoxiang Tian}, \bibinfo{person}{Guoquan Wu}, \bibinfo{person}{Jiren Yan}, \bibinfo{person}{Yan Jiang}, \bibinfo{person}{Jun Wei}, \bibinfo{person}{Wei Chen}, \bibinfo{person}{Shuo Li}, {and} \bibinfo{person}{Dan Ye}.} \bibinfo{year}{2022}\natexlab{b}.
\newblock \showarticletitle{Generating critical test scenarios for autonomous driving systems via influential behavior patterns}. In \bibinfo{booktitle}{\emph{Proceedings of the 37th IEEE/ACM International Conference on Automated Software Engineering}}. \bibinfo{pages}{1--12}.
\newblock


\bibitem[Vaswani et~al\mbox{.}(2017)]%
        {vaswani2017attention}
\bibfield{author}{\bibinfo{person}{Ashish Vaswani}, \bibinfo{person}{Noam Shazeer}, \bibinfo{person}{Niki Parmar}, \bibinfo{person}{Jakob Uszkoreit}, \bibinfo{person}{Llion Jones}, \bibinfo{person}{Aidan~N Gomez}, \bibinfo{person}{{\L}ukasz Kaiser}, {and} \bibinfo{person}{Illia Polosukhin}.} \bibinfo{year}{2017}\natexlab{}.
\newblock \showarticletitle{Attention is all you need}.
\newblock \bibinfo{journal}{\emph{Advances in neural information processing systems}}  \bibinfo{volume}{30} (\bibinfo{year}{2017}).
\newblock


\bibitem[Wang et~al\mbox{.}(2025)]%
        {wang2025moditector}
\bibfield{author}{\bibinfo{person}{Renzhi Wang}, \bibinfo{person}{Mingfei Cheng}, \bibinfo{person}{Xiaofei Xie}, \bibinfo{person}{Yuan Zhou}, {and} \bibinfo{person}{Lei Ma}.} \bibinfo{year}{2025}\natexlab{}.
\newblock \showarticletitle{MoDitector: Module-Directed Testing for Autonomous Driving Systems}.
\newblock \bibinfo{journal}{\emph{Proceedings of the ACM on Software Engineering}} \bibinfo{volume}{2}, \bibinfo{number}{ISSTA} (\bibinfo{year}{2025}), \bibinfo{pages}{137--158}.
\newblock


\bibitem[Watanabe et~al\mbox{.}(2018)]%
        {watanabe2018runtime}
\bibfield{author}{\bibinfo{person}{Kosuke Watanabe}, \bibinfo{person}{Eunsuk Kang}, \bibinfo{person}{Chung-Wei Lin}, {and} \bibinfo{person}{Shinichi Shiraishi}.} \bibinfo{year}{2018}\natexlab{}.
\newblock \showarticletitle{Runtime monitoring for safety of intelligent vehicles}. In \bibinfo{booktitle}{\emph{Proceedings of the 55th annual design automation conference}}. \bibinfo{pages}{1--6}.
\newblock


\bibitem[Wu et~al\mbox{.}(2024)]%
        {wu2024recent}
\bibfield{author}{\bibinfo{person}{Jingda Wu}, \bibinfo{person}{Chao Huang}, \bibinfo{person}{Hailong Huang}, \bibinfo{person}{Chen Lv}, \bibinfo{person}{Yuntong Wang}, {and} \bibinfo{person}{Fei-Yue Wang}.} \bibinfo{year}{2024}\natexlab{}.
\newblock \showarticletitle{Recent advances in reinforcement learning-based autonomous driving behavior planning: A survey}.
\newblock \bibinfo{journal}{\emph{Transportation Research Part C: Emerging Technologies}}  \bibinfo{volume}{164} (\bibinfo{year}{2024}), \bibinfo{pages}{104654}.
\newblock


\bibitem[Xu et~al\mbox{.}(2022)]%
        {xu2022safebench}
\bibfield{author}{\bibinfo{person}{Chejian Xu}, \bibinfo{person}{Wenhao Ding}, \bibinfo{person}{Weijie Lyu}, \bibinfo{person}{Zuxin Liu}, \bibinfo{person}{Shuai Wang}, \bibinfo{person}{Yihan He}, \bibinfo{person}{Hanjiang Hu}, \bibinfo{person}{Ding Zhao}, {and} \bibinfo{person}{Bo Li}.} \bibinfo{year}{2022}\natexlab{}.
\newblock \showarticletitle{Safebench: A benchmarking platform for safety evaluation of autonomous vehicles}.
\newblock \bibinfo{journal}{\emph{Advances in Neural Information Processing Systems}}  \bibinfo{volume}{35} (\bibinfo{year}{2022}), \bibinfo{pages}{25667--25682}.
\newblock


\bibitem[Zhang and Cai(2023)]%
        {zhang2023building}
\bibfield{author}{\bibinfo{person}{Xudong Zhang} {and} \bibinfo{person}{Yan Cai}.} \bibinfo{year}{2023}\natexlab{}.
\newblock \showarticletitle{Building Critical Testing Scenarios for Autonomous Driving from Real Accidents}. In \bibinfo{booktitle}{\emph{Proceedings of the 32nd ACM SIGSOFT International Symposium on Software Testing and Analysis}}. \bibinfo{pages}{462--474}.
\newblock


\bibitem[Zhang et~al\mbox{.}(2021)]%
        {roach_iccv}
\bibfield{author}{\bibinfo{person}{Zhejun Zhang}, \bibinfo{person}{Alexander Liniger}, \bibinfo{person}{Dengxin Dai}, \bibinfo{person}{Fisher Yu}, {and} \bibinfo{person}{Luc Van~Gool}.} \bibinfo{year}{2021}\natexlab{}.
\newblock \showarticletitle{End-to-end urban driving by imitating a reinforcement learning coach}. In \bibinfo{booktitle}{\emph{Proceedings of the IEEE/CVF international conference on computer vision}}. \bibinfo{pages}{15222--15232}.
\newblock


\bibitem[Zhao et~al\mbox{.}(2017)]%
        {zhao2017review}
\bibfield{author}{\bibinfo{person}{Zhiguo Zhao}, \bibinfo{person}{Liangjie Zhou}, \bibinfo{person}{Qiang Zhu}, \bibinfo{person}{Yugong Luo}, {and} \bibinfo{person}{Keqiang Li}.} \bibinfo{year}{2017}\natexlab{}.
\newblock \showarticletitle{A review of essential technologies for collision avoidance assistance systems}.
\newblock \bibinfo{journal}{\emph{Advances in Mechanical Engineering}} \bibinfo{volume}{9}, \bibinfo{number}{10} (\bibinfo{year}{2017}), \bibinfo{pages}{1687814017725246}.
\newblock


\bibitem[Zhong et~al\mbox{.}(2023)]%
        {tse_adfuzz}
\bibfield{author}{\bibinfo{person}{Ziyuan Zhong}, \bibinfo{person}{Gail Kaiser}, {and} \bibinfo{person}{Baishakhi Ray}.} \bibinfo{year}{2023}\natexlab{}.
\newblock \showarticletitle{Neural network guided evolutionary fuzzing for finding traffic violations of autonomous vehicles}.
\newblock \bibinfo{journal}{\emph{IEEE Transactions on Software Engineering}} \bibinfo{volume}{49}, \bibinfo{number}{4} (\bibinfo{year}{2023}), \bibinfo{pages}{1860--1875}.
\newblock


\bibitem[Zhou et~al\mbox{.}(2023)]%
        {zhou2023specification}
\bibfield{author}{\bibinfo{person}{Yuan Zhou}, \bibinfo{person}{Yang Sun}, \bibinfo{person}{Yun Tang}, \bibinfo{person}{Yuqi Chen}, \bibinfo{person}{Jun Sun}, \bibinfo{person}{Christopher~M Poskitt}, \bibinfo{person}{Yang Liu}, {and} \bibinfo{person}{Zijiang Yang}.} \bibinfo{year}{2023}\natexlab{}.
\newblock \showarticletitle{Specification-based Autonomous Driving System Testing}.
\newblock \bibinfo{journal}{\emph{IEEE Transactions on Software Engineering}} (\bibinfo{year}{2023}), \bibinfo{pages}{1--19}.
\newblock


\end{thebibliography}

\end{document}